\pdfoutput=1

\documentclass[11pt]{article}

\usepackage{acl}

\usepackage{stroop2}
\usepackage{times}
\usepackage{latexsym}

\usepackage[T1]{fontenc}

\usepackage[utf8]{inputenc}

\usepackage{microtype}

\usepackage{inconsolata}

\usepackage{graphicx}

\usepackage[dvipsnames]{xcolor}
\usepackage{hyperref}
\usepackage{inconsolata}
\usepackage{tabularx}
\usepackage{listings,lstautogobble}
\usepackage{fancyvrb}
\usepackage{fvextra}
\usepackage{tikz}
\usepackage{amsmath}
\usetikzlibrary{positioning}
\usepackage{pgf}
\usepackage{comment}
\usepackage{arydshln}
\usepackage{color} 
\usepackage{colortbl}
\usepackage{xcolor}
\usepackage{soul}

\newcommand{\towerinst}{TowerInstruct-13B}
\newcommand{\eurollm}{EuroLLM-9B-Inst}
\newcommand{\nllb}{NLLB-3.3B}
\newcommand{\gemmathree}{Gemma3-12B-it}
\newcommand{\qwenthree}{Qwen3-14B}
\newcommand{\BLEU}{BLEU}
\newcommand{\COMET}{COMET}
\newcommand{\CHRF}{ChrF}
\newcommand{\docCOMET}{docCOMET}
\newcommand{\COMETQE}{COMETQE}
\newcommand{\docCOMETQE}{docCOMETQE}
\newcommand{\enpt}{\textsc{EN-PT} }
\newcommand{\enge}{\textsc{EN-DE} }
\newcommand{\enfr}{\textsc{EN-FR} }
\newcommand{\enko}{\textsc{EN-KO} }
\newcommand{\enar}{\textsc{EN-AR} }
\newcommand{\enru}{\textsc{EN-RU} }

\newcommand{\newuline}[1]{{\color{black}\setulcolor{OliveGreen}\ul{#1}}}

\title{Unlocking Latent Discourse Translation in LLMs \\ Through Quality-Aware Decoding}

\author{Wafaa Mohammed$^1$  \qquad Vlad Niculae$^1$ \qquad Chrysoula Zerva$^{2,3}$  \\
$^1$University of Amsterdam \\
$^{2}$ Instituto Superior Técnico, University of Lisbon \\
$^{3}$Instituto de Telecomunicações   \\
  \texttt{\{w.m.a.mohammed, v.niculae\}@uva.nl}, \texttt{chrysoula.zerva@tecnico.ulisboa.pt}
}

\begin{document}
\maketitle
\begin{abstract}
Large language models (LLMs) have emerged as strong contenders in machine translation.Yet, they still struggle to adequately handle discourse phenomena, such as pronoun resolution and lexical cohesion at the document level. In this study, we thoroughly investigate the discourse phenomena performance of LLMs in context-aware translation. We demonstrate that discourse knowledge is encoded within LLMs and propose the use of quality-aware decoding (QAD) to effectively extract this knowledge, showcasing its superiority over other decoding approaches through comprehensive analysis. Furthermore, we illustrate that QAD enhances the semantic richness of translations and aligns them more closely with human preferences. \looseness-1
\end{abstract}

\section{Introduction}
Large language models (LLMs) have demonstrated superior performance in machine translation (MT), producing strong results for sentence-level and document-level translation \citep{wang-etal-2023-document-level,xu2023paradigm,DBLP:journals/corr/abs-2402-17733,zhu-etal-2024-multilingual}. Quality improvements in document-level translation are key in producing translations that align better with human preferences, since documents are the natural way in which we consume and produce text \citep{laubli-etal-2018-machine, DBLP:journals/csur/MarufSH21,mohammed-niculae-2024-measuring,dahan:hal-04798759}.
Additionally, document-level translation provides a means to tackle discourse-related challenges in translation, including inter-sentential coreference resolution as well as the need for maintaining coherence, style, and formality level across the document \citep{DBLP:journals/corr/abs-2304-12959}. 

\begin{table*}
\centering
\small
\begin{tabular}{l l}
\toprule
Lexical  &  \textbf{\textsc{EN:}} The \textcolor{purple}{\underline{reviewer}} gave us constructive feedback. We appreciate the \colorbox{pink}{reviewer}'s feedback.  \\ 
repetition & \textbf{\textsc{FR:}} L'\textcolor{purple}{\underline{examinatrice}} nous a fait un retour constructif. Nous apprécions le retour de l'\colorbox{pink}{examinatrice}.  \\
\midrule
Pronoun &  \textbf{\textsc{EN:}} One of the Chinese worked in an \textcolor{purple}{\underline{amusement park}}. \colorbox{pink}{It} was closed for the season.\\ 
resolution & \textbf{\textsc{DE:}} Ein Chinese arbeitete in einem \textcolor{purple}{\underline{Vergnügungspark}}. \colorbox{pink}{Er} war gerade geschlossen.\\
\midrule
\multirow{2}{*}{Formality} & \textbf{\textsc{EN:}} How are you my dear \textcolor{purple}{\underline{friend}}? Would \colorbox{pink}{you} like to go to the cinema with me?  \\ 
& \textbf{\textsc{DE:}} Wie geht es dir, mein lieber \textcolor{purple}{\underline{Freund}}? Möchtest \colorbox{pink}{du} mit mir ins Kino gehen? \\
\midrule
 \multirow{2}{*}{Verb form} &\textbf{\textsc{EN:}} \textcolor{purple}{\underline{Maria}} said she was too sick. However, she was \colorbox{pink}{seen} walking in the park. \\ 
& \textbf{\textsc{PT:}} A \textcolor{purple}{\underline{Maria}} disse que estava muito doente. No entanto, ela foi \colorbox{pink}{vista} a passear no parque. \\
\bottomrule
\end{tabular}
\caption{Examples of discourse phenomena. Ambiguous words are highlighted in \colorbox{pink}{pink}, and supporting context necessary to resolve the ambiguity is marked in \textcolor{purple}{\underline{underlined purple}} text.}
\label{phenomena_examples}
\end{table*}

At the same time, it has been observed that LLM-derived translations frequently feature different linguistic and semantic characteristics and patterns compared to neural machine translation (NMT), hence inspiring several works that try to trace and understand such patterns and differences. 
Thus, recent work ranges from designing linguistic performance test suites \citep{manakhimova-etal-2024-investigating} to analyzing specific aspects such as lexical features, literalness, formality \citep{wisniewski-etal-2024-fame}, gender bias \citep{DBLP:conf/ci2/KotekDS23,DBLP:journals/corr/abs-2403-00277}, and pronoun resolution. These studies uncovered valuable features of LLM-derived translations, including suboptimal performance compared to NMT systems in several phenomena, such as punctuation, future verb tenses, stripping, function words \citep{manakhimova-etal-2024-investigating}, and pronoun resolution \citep{mohammed2024analyzing}. Other works observed that LLMs show systematic differences to NMT systems in their choice of lexical features, such as Part-of-speech (PoS) patterns \citep{sizov-etal-2024-analysing} as well as their ability to produce less literal translations while remaining competitive quality-wise to NMT translations \citep{raunak-etal-2023-gpts}. \looseness-1

Despite these insights, fine-grained analyses rarely extend to document-level MT, where discourse context makes such phenomena even more critical and further underscores the need to understand the linguistic and semantic properties of LLM translations. We thus aim to study the performance of LLMs in document-level MT (particularly, context-aware MT) with respect to different discourse phenomena. Inspired by \citet{fernandes-etal-2023-translation}, we measure models' performance on four phenomena: lexical repetition, pronoun resolution, formality, and verb forms. We compare the performance of recent translation-LLMs to encoder-decoder models on the DELA corpus, a high-quality human-curated dataset that is rich in discourse phenomena \citep{castilho-etal-2021-dela}. Moreover, we hypothesize that discourse knowledge can be implicitly encoded in LLMs, but is not fully exploited by greedy decoding. We thus experiment with quality-aware decoding \citep{fernandes-etal-2022-quality} and find that it indeed helps improve the discourse phenomena performance of LLMs. We validate our findings through extensive experiments on six language pairs from three language families: English to Brazilian-Portuguese, German, French, Korean, Arabic and Russian, on two datasets, namely, TED2020 \citep{reimers-gurevych-2020-making} and WMT24++ dataset \citep{DBLP:journals/corr/abs-2502-12404}. Moreover, we perform an ablation study on different inference setups, including MBR decoding \citep{eikema-aziz-2020-map} with different choices for the utility function incorporating discourse specific metrics \citep{wong-kit-2012-extending,zhao-etal-2023-discoscore}, automatic post editing, and sample fusion \citep{vernikos-popescu-belis-2024-dont}.

Our contributions can be summarized as follows:
\begin{itemize}[topsep=0pt, partopsep=0pt, parsep=0pt, itemsep=0pt]
    \item We design a comprehensive evaluation setup leveraging a discourse-rich dataset, showing that under greedy decoding, encoder-decoder models outperform LLMs in terms of discourse performance.\looseness-1
    \item We demonstrate through extensive evaluation on six language pairs using automatic metrics, LLM-as-a-judge, and human assessment that QAD improves the translation and the discourse performance of LLMs, enabling them to surpass encoder-decoders. 
    \item We conduct a comprehensive analysis on the effect of different inference setups on discourse performance.\looseness-1
    \item We release human annotations based on TED2020 that focus on discourse phenomena, supporting further research in this area.\footnote{All code and data will be released upon acceptance.}
\end{itemize}

\section{Background}
\subsection{Discourse Phenomena in Document-Level Translation}
Translating beyond the sentence level brings extra challenges that concern inter-sentential coreference resolution, lexical repetition, and coherence. Handling these challenges is important to ensure reliable, adequate translations that align with human preferences. In this work, we focus on four linguistic phenomena that are relevant to document-level translation as proposed by \citep{fernandes-etal-2023-translation}:
\textbf{Lexical repetition.} Entities mentioned multiple times in a document should be translated in the same way.\\
\textbf{Pronoun resolution.} For languages that have gendered pronouns, the translation should respect the gender of the referent.\\
\textbf{Formality.} In some languages, linguistic indicators such as pronouns and honorifics are used when addressing someone formally or expressing respect. The same level of formality should be maintained across a document. \\
\textbf{Verb form.} In some languages, verb forms vary depending on the subject, tense, and consistency with the context. For instance, in Arabic, the verb form changes according to whether the subject is singular, dual, or plural, masculine or feminine. Verb forms should be accurate across a document.

Examples of the phenomena are presented in \Cref{phenomena_examples}. The discourse phenomena we study here are for EN$\rightarrow$XX translation direction. Studying the reverse direction (XX $\rightarrow$EN) requires a separate set of analyses to determine the relevant phenomena in that case, which is out of scope of this work.

\subsection{Quality-Aware Decoding (QAD)}
Quality-aware decoding for machine translation refers to utilizing translation evaluation metrics during decoding to choose the best candidate among several sampled responses from the model. Samples can be generated using vanilla temperature sampling or variations of it that truncate the distribution, such as top-k or nucleus sampling \citep{fan-etal-2018-hierarchical,DBLP:conf/iclr/HoltzmanBDFC20}. QAD has been proven to generate better quality translations compared to maximum-a-posteriori (MAP) decoding according to automatic metrics and human evaluation \citep{fernandes-etal-2022-quality}. There are different approaches to quality aware-decoding including reranking \citep{lee-etal-2021-discriminative,bhattacharyya-etal-2021-energy}, minimum Bayes risk (MBR) decoding \citep{eikema-aziz-2020-map,eikema-aziz-2022-sampling,muller-sennrich-2021-understanding}, and fusion of samples \citep{vernikos-popescu-belis-2024-dont}. In our main experiments, we focus on MBR since it is a widely used approach for MT with proven benefits on overall translation performance \citep{wu-etal-2024-choose,li-etal-2024-scir,kudo-etal-2024-document}. We explore whether those improvements are reflected in discourse performance.\\

\noindent
\textbf{MAP.} A MT model defines a probability distribution $p(y|x,\theta)$ over a set of hypotheses $\mathcal{Y}$. MAP decoding, such as greedy decoding, aims to maximize the probability of generated hypothesis: 
\begin{equation}
    \hat{h} = \underset{y \in \mathcal{Y}} {\argmax} \: p(y|x,\theta).
\end{equation}
\textbf{MBR.} Given a utility function $u$ that measures the similarity between a hypothesis $h$ and a reference $y$, MBR decoding aims to find the hypothesis that maximizes the expected utility (minimizes the loss) among a set of sampled hypotheses $\mathcal{H}$ (each sampled hypothesis is used as a pseudo reference and compared to all other hypotheses). It selects:
\begin{equation}
    \hat{h} = \underset{h \in \mathcal{H}}{\argmax} \, \bbE_{y \sim p(y|x,\theta)} \: [u(h,y)]. 
\end{equation}

We experiment with different choices of utility functions, including lexical, pretrained, and discourse-specific metrics for translation evaluation. Additionally, we experiment with other inference approaches, including fusion and automatic post-editing. We discuss these in detail in \cref{inference-setup}. 

\begin{table*}
\centering
\small
\setlength{\tabcolsep}{4pt}
\begin{tabular}{lcccccccccc}
\toprule
& \BLEU & \COMET & \docCOMET & \COMETQE & \docCOMETQE & L.repetition & Formality & Pronouns \\
\midrule
\textbf{ctx= 0} \\
\textcolor{gray}{nllb G} & \textcolor{gray}{55.2} & \textcolor{gray}{87.1} & \textcolor{gray}{82.2} & \textcolor{gray}{81.5} & \textcolor{gray}{75.4} & \textcolor{gray}{\textbf{87.0}} & \textcolor{gray}{\textbf{75.0}}& \textcolor{gray}{45.0} \\
\textcolor{gray}{nllb Q} & \textcolor{gray}{\textbf{58.2}} & \textcolor{gray}{87.4} & \textcolor{gray}{82.3} & \textcolor{gray}{81.6} & \textcolor{gray}{76.5} & \textcolor{gray}{85.0} \textbf{$\color{red}{\downarrow}$} & \textcolor{gray}{\textbf{76.0}} & \textcolor{gray}{47.0} \\
tower G & 34.4 & 84.7 & 79.1 & {79.7} & {74.6} & 77.0 & 51.0 & 32.0 \\
tower Q & 52.7 & 88.9 & 84.3 & {\textbf{83.0}} & {\textbf{80.1}} & 85.0 & 68.0 & 39.0 \\
euro G & {30.7} & 82.5 & 75.2 & 77.9 & 70.1 & 78.0 & {59.0} & 33.0 \\
euro Q & {52.4} & {87.9} & 81.5 & {82.5} & {79.8} & 86.0 & \textbf{75.0} & 45.0 \\
gemma G & 50.6 & 85.6 & 80.1 & 79.4 & 75.9 & 82.0 & 12.0 & 38.0 \\
gemma Q & 53.0 & 88.5 & 83.5 & {82.3} & {78.8} & 84.0 & {72.0} & 41.0 \\
qwen G & 50.9 & 88.5 & 83.6 & {82.1} & 78.9 & 84.0 & \textbf{73.0} & 41.0 \\
qwen Q & {54.1} & {{88.9}} & {84.0} & 82.4 & {79.3} & 84.0 & \textbf{74.0} & 40.0 \textbf{$\color{red}{\downarrow}$} \\
\midrule
\textbf{ctx= 5} \\
tower G & \newuline{41.0} & \newuline{86.0} & \newuline{81.2} & 79.1 & 74.5 & \newuline{85.0} & \newuline{66.0} & \newuline{50.0} \\
tower Q & \newuline{57.4} & \newuline{\textbf{89.6}} & \newuline{\textbf{85.4}} & 82.0 & 79.5 & \newuline{\textbf{90.0}} & \newuline{\textbf{76.0}} & \newuline{\textbf{60.0}} \\
euro G & 25.9 & {80.4} & {73.6} & {76.0} & {66.1} & \newuline{79.0} & 56.0 & \newuline{40.0} \\
euro Q & 52.1 & 87.8 & \newuline{82.0} & 81.9 & 78.9 & \newuline{\textbf{89.0}} & \newuline{\textbf{75.0}} & \newuline{48.0} \\
gemma G & \newuline{54.0} & \newuline{87.3} & \newuline{82.3} & \newuline{80.3} & \newuline{77.4} & \newuline{86.0} & \newuline{27.0} & \newuline{\textbf{49.0}} \\
gemma Q & \newuline{56.2} & \newuline{88.5} & \newuline{83.8} & 81.5 & 78.5 & \newuline{87.0} & 51.0 & \newuline{\textbf{52.0}} \\
qwen G & \newuline{52.4} & \newuline{89.3} & \newuline{84.9} & 82.0 & \newuline{79.3} & \newuline{\textbf{89.0}} & \newuline{\textbf{75.0}} & \newuline{\textbf{51.0}} \\
qwen Q & 52.4 & \newuline{\textbf{89.6}} & \newuline{85.3} & {82.2} & \newuline{\textbf{79.7}} & \newuline{\textbf{90.0}} & \newuline{\textbf{76.0}} & \newuline{\textbf{52.0}} \\
\bottomrule
\end{tabular}
\caption{Translation and discourse phenomena performance of all models (nllb={\nllb}, tower={\towerinst}, euro={\eurollm}, gemma={\gemmathree}, qwen={\qwenthree}) using greedy (G) and quality-aware decoding (Q) on \textbf{DELA dataset} in both sentence-level (ctx= 0) and context-aware (ctx= 5) setups. \textbf{Bold} highlights the best value per column (with statistical significance p<0.05). The numbers presented for phenomena are F1 accuracies (details in \Cref{muda-description}). Cases where the QAD result is worse than its Greedy counterpart are marked with a red down arrow \textbf{$\color{red}{\downarrow}$}. Cases where the context-aware result is better than the sentence level result are marked with a \newuline{green underline}.}
\label{dela_table}
\end{table*}

\section{Experiments\footnote{Sustainability statement for all experiments: \Cref{sustainability}.}}

\subsection{Data}
We experiment on the DELA corpus \citep{castilho-etal-2021-dela}, an English-Brazilian-Portuguese document-level corpus annotated with context-related issues. The corpus is a collection of 60 documents with 3.7K sentences from different domains (news, subtitles, literature, legislation, reviews, medical) that are manually selected, translated, and annotated with context-dependent discourse phenomena. Additionally, we experiment on a subset of TED2020 data (20K sentences in around 160 documents) \citep{reimers-gurevych-2020-making} available in OPUS \citep{tiedemann-2012-parallel}. We also experiment on the WMT24++ dataset \citep{DBLP:journals/corr/abs-2502-12404}, which has a size of approximately 1K sentences across 169 documents (results are in \Cref{wmt24_results_appendix}). For both TED2020 and WMT24++, we experiment on six language directions: English (\textsc{EN}) to Brazilian-Portuguese (\textsc{PT}), German (\textsc{DE}), French (\textsc{FR}), Korean (\textsc{KO}), Arabic (\textsc{AR}) and Russian (\textsc{RU}). Dataset statistics for the three corpora, including discourse phenomena statistics, are presented in \Cref{dataset_statistics_appendix}.

\subsection{Models}
We prioritize models with high performance in multilingual and MT-related tasks. Thus, we evaluate {\towerinst} \citep{DBLP:journals/corr/abs-2402-17733}, an instruction-tuned translation-LLM based on Llama2-13B \citep{DBLP:journals/corr/abs-2307-09288}, which is one of the leading translation systems according to WMT24 \citep{kocmi-etal-2024-findings}. We chose the best-performing version of the model within our budget. We also evaluate {\eurollm} \citep{DBLP:journals/corr/abs-2409-16235}, a leading European LLM\footnote{based on the european-llm-leaderboard \href{https://huggingface.co/spaces/Eurolingua/european-llm-leaderboard}{https://huggingface.co/spaces/Eurolingua/european-llm-leaderboard}} trained from scratch on all European Union languages and additional relevant ones. {\eurollm} is trained on all translation data used for {\towerinst}, as well as additional sources, giving it broader coverage.\looseness-1 

For generalizability, we include LLMs with strong performance on general language generation tasks, namely {\gemmathree} \citep{DBLP:journals/corr/abs-2503-19786} and {\qwenthree} \citep{DBLP:journals/corr/abs-2505-09388}.
As an encoder-decoder baseline, we include {\nllb} \citep{DBLP:journals/corr/abs-2207-04672}, a leading sentence-level multilingual NMT model with strong performance across languages of interest in this work. Context-aware encoder-decoder variants (see \Cref{related-work}) would also be worth exploring, but have limited language coverage and are beyond the scope of this work.\looseness-1

\subsection{Inference}
\label{inference}
We compare two decoding setups: \textbf{greedy} decoding, which selects the highest-probability token at each step, and quality-aware decoding (\textbf{QAD}), which uses MBR with 50 samples generated via nucleus sampling (p=0.9)\footnote{For {\qwenthree}, we use the non-thinking mode with the sampling parameters recommended by its developers: t=0.7, topP=0.8, topK=20, minP=0. \href{https://huggingface.co/Qwen/Qwen3-14B}{https://huggingface.co/Qwen/Qwen3-14B}}. We use \textbf{BLEU} score \citep{DBLP:conf/acl/PapineniRWZ02} as utility function for all our experiments unless otherwise indicated.
Note that we first conducted preliminary experiments on different prompting formats for each model and present only the best setup in this work. For LLMs (all models except {\nllb}), we employ context-aware prompting with the context being (up to) 5 previous source-target pairs in the same document (prompt formats in \Cref{prompt-formats-appendix}). We opt for the context-aware setup instead of translating documents holistically to eliminate the potential bias of how models handle different parts of the context \citep{liu-etal-2024-lost}.
For {\nllb}, since the model has only been trained on sentence-level data, we conduct inference at the sentence level. We also report sentence-level baseline results for LLMs.\looseness-1

\begin{table*}
\centering
\small
\setlength{\tabcolsep}{4pt}
\begin{tabular}{lcccccccccc}
\toprule
& \BLEU & \COMET & \docCOMET & \COMETQE & \docCOMETQE & L.repetition & Formality & Pronouns \\
\midrule
\textbf{ctx= 0} \\
\textcolor{gray}{nllb G} & \textcolor{gray}{28.3} & \textcolor{gray}{84.2} & \textcolor{gray}{79.2} & \textcolor{gray}{82.7} & \textcolor{gray}{80.0} & \textcolor{gray}{64.2} & \textcolor{gray}{57.4} & \textcolor{gray}{61.2} \\
\textcolor{gray}{nllb Q} & \textcolor{gray}{29.3} & \textcolor{gray}{84.3} & \textcolor{gray}{79.4} & \textcolor{gray}{82.6} \textbf{$\color{red}\downarrow$} & \textcolor{gray}{80.0} & \textcolor{gray}{64.7} & \textcolor{gray}{57.6} & \textcolor{gray}{60.5} \textbf{$\color{red}\downarrow$} \\
tower G & 21.0 & {81.5} & {76.1} & {80.1} & {75.1} & 60.0 & 48.8 & 51.3 \\
tower Q & 31.1 & 85.7 & 80.9 & {\textbf{84.2}} & {\textbf{81.9}} & 66.4 & 58.2 & 60.3 \\
euro G & {15.8} & {79.6} & {73.0} & {78.3} & {71.0} & {56.8} & 48.0 & 50.0 \\
euro Q & 26.6 & 84.7 & 79.0 & {83.5} & {81.1} & 63.8 & 59.0 & 59.0 \\
gemma G & 25.1 &  {81.9} & {76.2} & {80.5} & {78.3} & 60.3 & 16.4 & 55.2 \\
gemma Q & 26.4 & {84.4} & {79.1} & {83.2} & {80.7} & 62.5 & {55.4} &  57.5 \\
qwen G & 24.8 &  83.4 & 78.1 & 82.2 & 79.0 & 61.8 & 50.4 & 57.8 \\
qwen Q & 25.9 & 84.4 & 79.1 & 83.2 & 80.8 & 63.3 & 52.2 & 59.0 \\
\midrule
\textbf{ctx= 5} \\
tower G & \newuline{21.0} & 79.6 & 74.8 & 75.1 & 71.4 & \newuline{62.0} & \newuline{52.2} & \newuline{62.0} \\
tower Q & \newuline{\textbf{33.0}} & \newuline{\textbf{86.1}} & \newuline{\textbf{82.0}} & 82.5 & 80.1 & \newuline{\textbf{71.2}} & \newuline{\textbf{64.6}} & \newuline{71.0} \\
euro G & 15.0 & 78.7 & 72.7 & 77.0 & 68.9 & \newuline{60.5} & \newuline{49.4} & \newuline{52.8} \\
euro Q & \newuline{28.2} & \newuline{85.1} & \newuline{79.9} & 83.2 & 80.8 & \newuline{67.7} & \newuline{62.4} & \newuline{63.0} \\
gemma G & \newuline{26.9} & 80.9 & 75.4 & 79.1 & 76.6 & \newuline{63.2} & \newuline{18.4} & \newuline{58.8} \\
gemma Q & \newuline{27.8} & 83.8 & 78.5 & 82.3 & 80.0 & \newuline{64.8} & 42.6 & \newuline{58.8} \\
qwen G & \newuline{26.7} & \newuline{84.4} & \newuline{79.4} & \newuline{82.3} & \newuline{79.5} & \newuline{66.7} & \newuline{61.0} & \newuline{61.2}  \\
qwen Q & \newuline{27.9} & \newuline{85.2} & \newuline{80.4} & \newuline{83.2} & \newuline{81.1} & \newuline{67.0} & \newuline{62.0} & \newuline{\textbf{82.8}} \\
\bottomrule
\end{tabular}
\caption{Translation and discourse phenomena performance of all models (nllb={\nllb}, tower={\towerinst}, euro={\eurollm}, gemma={\gemmathree}, qwen={\qwenthree}) using greedy (G) and quality-aware decoding (Q) on \textbf{TED2020 dataset} in both sentence-level (ctx= 0) and context-aware (ctx= 5) setups. The results are averaged across all language pairs. \textbf{Bold} highlights the best value per column. Cases where the QAD result is worse than its Greedy counterpart are marked with a red down arrow \textbf{$\color{red}{\downarrow}$}. Cases where the context-aware result is better than the sentence level result are marked with a \newuline{green underline}.}
\label{averaged_ted_results}
\end{table*}

\subsection{Evaluation}

\subsubsection{Overall Translation Evaluation}
We use a lexical metric, {\BLEU}\footnote{SacreBLEU signatures in \Cref{metric-signatures}} \citep{DBLP:conf/acl/PapineniRWZ02}, a reference-based pretrained metric, {\COMET}\footnote{\href{https://huggingface.co/Unbabel/wmt22-comet-da}{https://huggingface.co/Unbabel/wmt22-comet-da}} and its document-level variant {\docCOMET} \citep{rei-etal-2022-comet}, and a reference-free pretrained metric, {\COMETQE}\footnote{\href{https://huggingface.co/Unbabel/wmt22-cometkiwi-da}{https://huggingface.co/Unbabel/wmt22-cometkiwi-da}} \citep{rei-etal-2022-cometkiwi} and its document-level variant {\docCOMETQE}. \looseness-1
\subsubsection{Discourse Phenomena Evaluation}
\label{muda-description}
We measure F1 accuracy of tagged words with discourse phenomena in the reference being tagged in the hypothesis. To do so, we utilize the multilingual discourse-aware benchmark (MuDA) for discourse phenomena evaluation \citep{fernandes-etal-2023-translation}. Tagging is done automatically using predefined language-specific lists of pronouns, verb forms, and formality indicators. For lexical repetition, tagging is performed by obtaining source-target word alignments; if an alignment pair occurs more than a specific number of times (three in our experiments, following MuDA), the word is tagged for lexical repetition.

\subsubsection{LLM-Based Evaluation}
Evaluating LLMs automatically has become increasingly difficult due to their rapid advancements. Consequently, using language models to automatically assess long-form text (LLM-as-a-judge) is gaining popularity. We employ the multilingual M-Prometheus \citep{pombal2025mprometheussuiteopenmultilingual} judge in an absolute evaluation setup where the judge is provided with the instruction used to prompt the translation model, along with the translation output. The judge assigns a rating between 1 and 5, accompanied by an explanation of the decision. Since we use different prompting setups (\cref{inference}), the instructions provided to the judge are different, which makes direct comparisons unfair. Therefore, we report only the difference between greedy and QAD scores for each model rather than their absolute scores. \looseness-1

\section{Results}
\begin{figure}[t]
\centering
\includegraphics[width=\columnwidth]{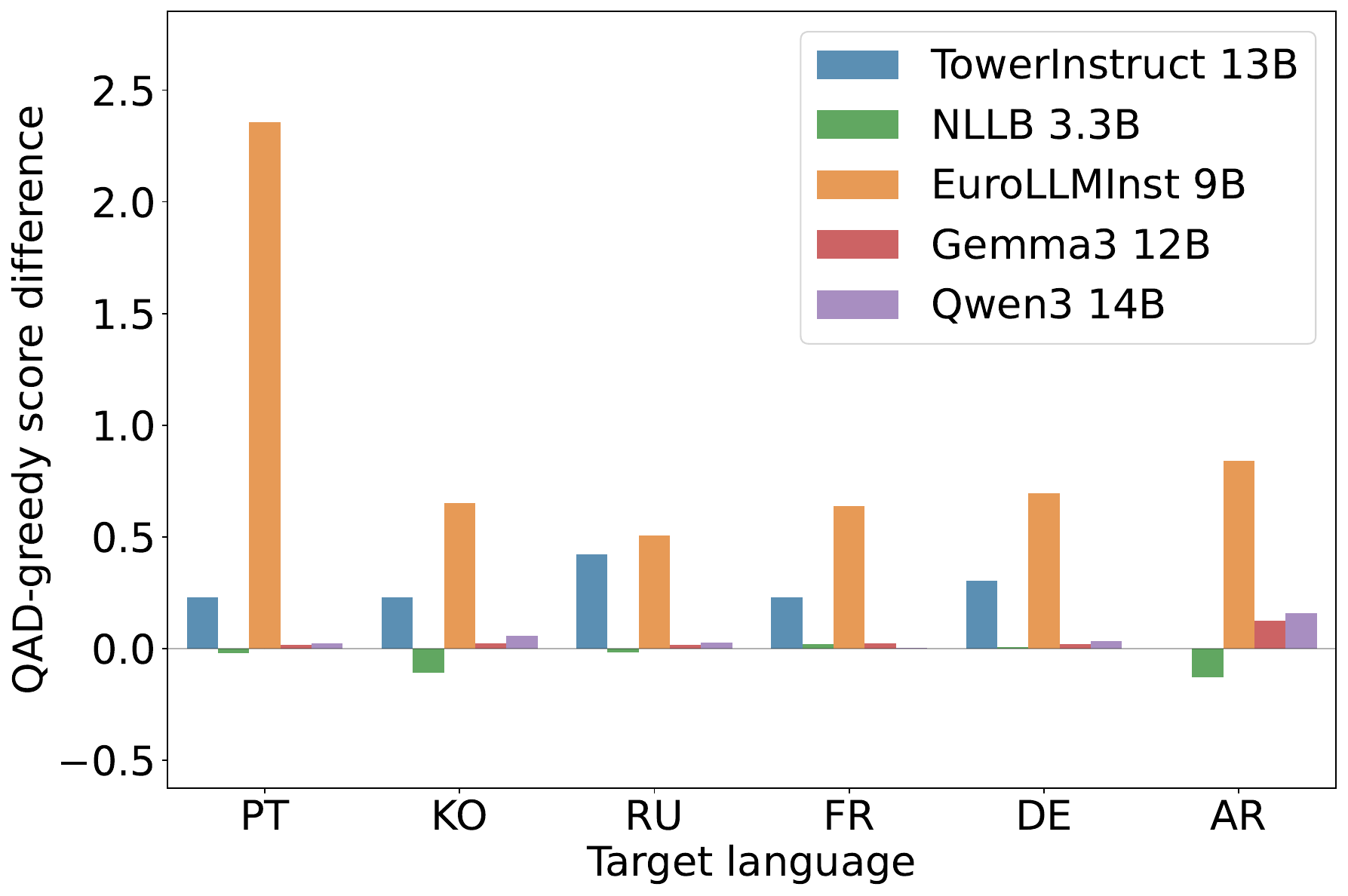}
\caption{Difference between QAD and greedy LLM-as-a-judge scores on TED2020 data. The plot demonstrates that QAD improves the performance of LLMs.} \label{llm-as-a-judge-plot}
\end{figure}

In \Cref{dela_table}, we present the results on the DELA corpus. We see that LLMs often fall behind {\nllb} in translation and discourse phenomena performance when using greedy decoding. Interestingly, we observe that using QAD notably improves both overall translation and discourse phenomena handling of LLMs, allowing them to outperform {\nllb}\footnote{While NLLB performs well on discourse phenomena, it lacks a reliable solution: without access to context, it guesses outputs that may align with the intended translation but remain inconsistent and unprincipled.}. 
In \Cref{averaged_ted_results} we show the results on the TED2020 dataset averaged across all language pairs. WMT24++ results are deferred to \Cref{wmt24_results_appendix} as they evidence similar overall trends. Detailed language-specific results are in \Cref{appendix-language-specific}.
The results highlight the substantial improvements in discourse and translation performance of translation-finetuned LLMs ({\towerinst} and {\eurollm}) using QAD. QAD also improves the performance of general-purpose LLMs ({\qwenthree} and {\gemmathree}), though the improvements are more modest; this could be attributed to the lack of translation-specific specialization in those models. Among all languages and datasets tested, {\towerinst} achieves the best overall performance, highlighting the effectiveness of specialized translation finetuning in encoding discourse awareness in LLMs.
Additionally, we present the differences between greedy and QAD scores from the LLM-as-a-judge evaluation for TED2020 data in \Cref{llm-as-a-judge-plot}, WMT24++ results are in \Cref{wmt24_results_appendix}. The results, consistent with automatic metrics, show that QAD noticeably enhances the performance of (\towerinst, \eurollm), while the improvements for {\gemmathree} and {\qwenthree} are relatively modest. In contrast, QAD does not improve on the performance of {\nllb}. 

From \Cref{dela_table,averaged_ted_results}, the effectiveness of QAD is evident in both sentence-level and context-aware setups of LLMs. Although sentence-level setups sometimes achieve higher overall translation performance (specially with greedy decoding), context-aware setups consistently achieve better performance on discourse phenomena. Notably, {\COMETQE} and {\docCOMETQE} tend to rate sentence-level outputs higher than context-aware ones, unlike other metrics. Through further analysis and manual inspection, we find that while fluency and correctness are comparable, sentence-level outputs are shorter on average, suggesting a bias in {\COMETQE} toward brevity. Similar bias toward shorter tanslations has been observed for other COMET variants (e.g. XCOMET; \citet{guerreiro-etal-2024-xcomet}) in \citet{perrella-etal-2024-guardians}. In combination with our findings, we hope it motivates further research on the behavior and robustness of such metrics for context-aware translation evaluation.

\begin{table*}
\centering
\small
\setlength{\tabcolsep}{4pt}
\begin{tabular}{l l l l l l l l l l}
\toprule
 & \BLEU & \CHRF & \COMET & \docCOMET & \COMETQE & \docCOMETQE & Rep. & Formal. & Pro. \\
\midrule
Greedy & 41.0 & 66.6 & 86.0 & 81.2 & 79.1 & 74.5 & 85 & 66 & 50 \\
QAD(\BLEU) & {\textbf{57.4}}* & 76.3* & 89.6* & 85.4* & 82.0* & 79.5* &\textbf{90}* & \textbf{76}* & \textbf{60}* \\
QAD(\CHRF) & 55.8* & {\textbf{76.9}}* & 89.7* & 85.4* & 82.2* & 79.6* &  \textbf{90}* & \textbf{77}* & \textbf{61}* \\
QAD(\COMET) & 54.2* & 75.0* & {\textbf{90.9}}* & {\textbf{86.2}}* & 83.1* & 80.7* & \textbf{89}* & \textbf{76}* & \textbf{62}* \\
QAD(LC) & 41.3 & 67.5* & 84.6*  & 81.8* & 79.6* & 75.5*  & 85 & 64 & 49 \\
QAD(Disco) & 55.3*  & 75.1* & 89.4* & 85.0* & 81.8* & 78.9*  & \textbf{89}* & \textbf{76}* & \textbf{57}* \\
Fusion & 41.6 & 67.8* & 89.1*  &  84.1* & {\textbf{85.7}}* & {\textbf{82.5}}* &  86 & 67 & 46 \\
APE & 44.3* & 68.4* & 87.7* & 82.6* & 82.0* & 77.9* & 84 & 69 & 47 \\
\bottomrule
\end{tabular}
\caption{Translation and discourse phenomena performance of different decoding setups using {\towerinst} on DELA data. \textbf{Bold} highlights the best value per column (with statistical significance p<0.05). Statistically significant outputs compared to the greedy output are marked with "*" (paired bootstrap resampling \citep{koehn-2004-statistical}). 
}
\label{inference-ablation-table}
\end{table*}
\begin{table}
\centering
\begin{tabular}{ l c}
\toprule
 Setup & Edit rate\\
\midrule
QAD (\BLEU) &  32.7\\
QAD (\CHRF) & 33.6 \\
QAD (\COMET) & 35.4 \\
QAD (LC) & 46.4 \\
QAD (Disco) & 34.5 \\
Fusion & 45.1 \\
APE & 29.6 \\
\bottomrule
\end{tabular}
\caption{Edit rate against the greedy output.}
\label{edit_rate_table}
\end{table}
\section{Analysis}
\subsection{Inference Setup Ablation}
\label{inference-setup}
We perform an ablation study on the entire DELA data using the {\towerinst} model, comparing different inference setups. Specifically:
\paragraph{QAD.} We explore the following utility functions:
\begin{itemize}[topsep=0pt, partopsep=0pt, parsep=0pt, itemsep=0pt]
    \item \textbf{Translation metrics.} {\BLEU}, {\CHRF} \citep{popovic-2015-chrf} and {\COMET} scores. Here, we use MBR with 50 samples-per-instance generated using nucleus sampling (p=0.9).
    \item \textbf{Discourse-specific metrics.} Lexical cohesion (LC) ratio \citep{wong-kit-2012-extending}, which is the number of lexical cohesion devices (repetitions, hypernyms, and synonyms) divided by the total number of content words, and DiscoScore \citep{zhao-etal-2023-discoscore}, a parametrized metric that uses BERT \citep{devlin-etal-2019-bert} to model discourse coherence through sentence graphs. Here, we use MBR with 20 samples-per-instance generated using nucleus sampling (p=0.9). \footnote{We use 20 samples instead of 50 due to computational constraints, as the discourse metrics involve generating an entity graph for each sample, which becomes impractical with a higher number of samples.}
\end{itemize}

\paragraph{Fusion.} Proposed by \citet{vernikos-popescu-belis-2024-dont}, the approach works by combining spans from different candidates generated via nucleus sampling (p=0.9) using a QE metric (\COMETQE).

\paragraph{Automatic post editing (APE).} Editing greedy outputs using XTOWER \citep{treviso-etal-2024-xtower} and XCOMET \citep{guerreiro-etal-2024-xcomet}, as in the IT-Unbabel team's submission to the quality estimation shared task at WMT24 \citep{zerva-etal-2024-findings}.

We assess the methods based on translation and discourse phenomena performance. Based on \Cref{inference-ablation-table}, QAD outperforms other inference approaches, including fusion and APE. Notably, translation metrics are more effective utility functions compared to discourse-specific metrics, with lexical measures (\BLEU, \CHRF) slightly outperforming the pretrained {\COMET}, though overall performance remains comparable.
Statistical significance testing, using greedy decoding as the baseline, shows significant improvements across all metrics for QAD with \BLEU, \CHRF, \COMET, and isco. However, QAD (LC), fusion, and APE are not outperforming the greedy baseline for discourse phenomena.\looseness-1

Inspired by \citet{zerva-etal-2024-findings}, to gain deeper insight into the discourse-related revisions introduced by each setup, we analyze the edit rate in their outputs relative to greedy decoding, using it as a \textit{proxy} for discourse-related revisions. The analysis focuses on sentences tagged with discourse phenomena using MuDA \citep{fernandes-etal-2023-translation}. 
In \Cref{edit_rate_table}, we present the edit rate of each setup compared to the greedy output. Looking at \Cref{inference-ablation-table,edit_rate_table}, we observe that the edit rate is aligned with discourse and translation performance. It seems that a moderate level of edit operations (insert, delete, substitute, shift) produces strong translation and discourse performance results, as demonstrated by the utility functions \BLEU, \CHRF, \COMET, and DiscoScore. However, deviations from this balance, whether through fewer edits (APE) or excessive edits (LC and fusion), compromise performance. Overall, this analysis highlights that among the experimented setups, \textbf{QAD with translation metrics is the best setup to improve discourse performance}.

\begin{figure}[t]
\centering
\includegraphics[width=\columnwidth]{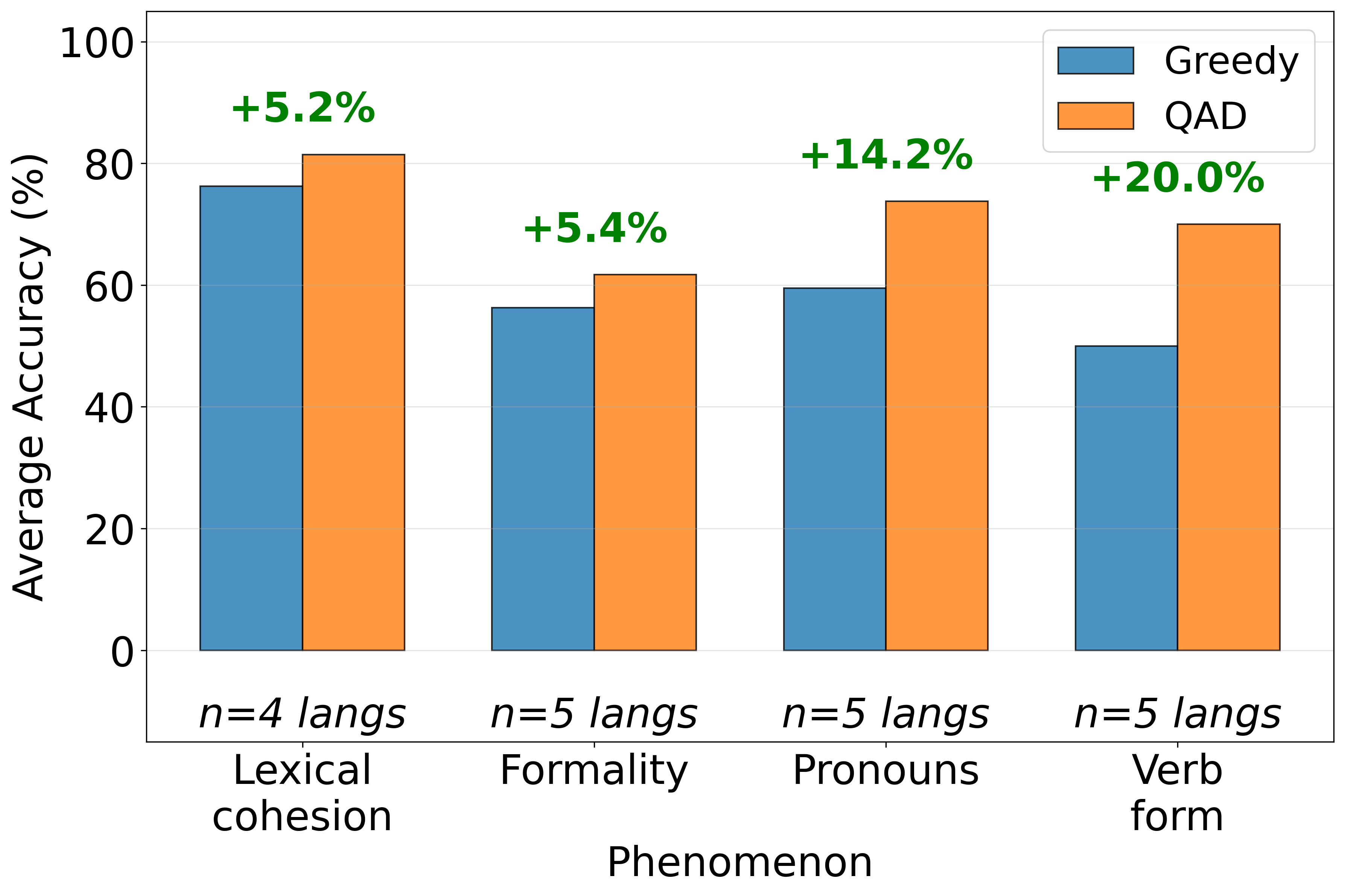}
\caption{Human-annotated accuracy of greedy and QAD outputs in handling discourse phenomena, averaged across all languages where the phenomena occur (number of languages is at the bottom). Arabic is excluded from this plot to avoid model-specific biases.}
\label{performance-plot}
\end{figure}
\begin{figure}[t]
\centering
\includegraphics[width=\columnwidth]{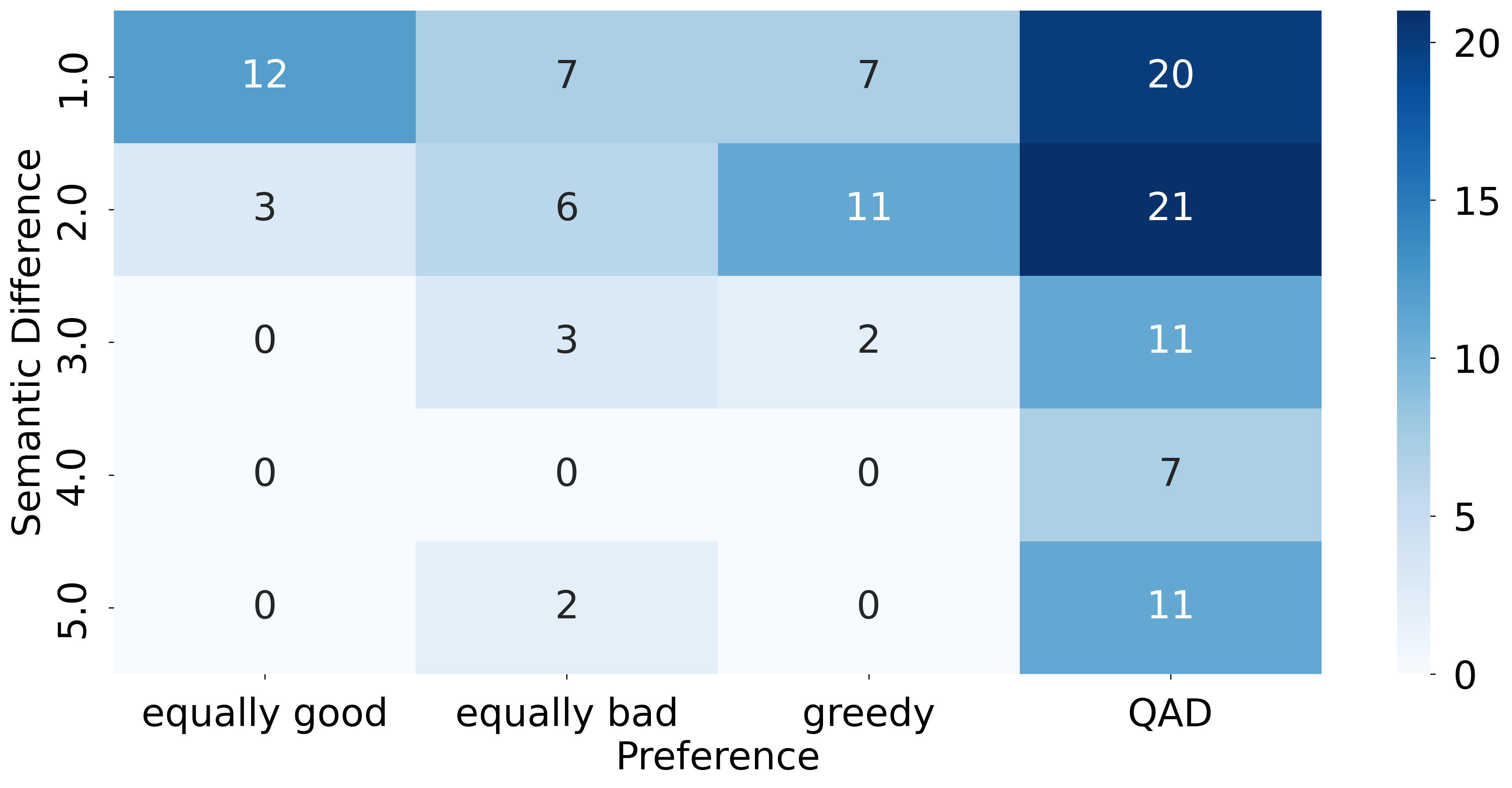}
\caption{Semantic difference vs. preference summed over all languages (except Arabic).} \label{preference-plot}
\end{figure}

\subsection{Human Evaluation and Qualitative Analysis}
We conduct a small-scale manual evaluation of the outputs to examine the semantic differences between greedy and QAD outputs, and to confirm the findings of the automated evaluation of discourse phenomena, which relies on MuDA \citep{fernandes-etal-2023-translation}.
We use the outputs of the best-performing model on TED2020 data, which is {\towerinst} for all languages except Arabic, where we use {\eurollm}. We randomly sample a subset of 25 instances of \texttt{\{source, greedy\_MT, QAD\_MT\}} for each language, all annotated with discourse phenomena via MuDA \citep{fernandes-etal-2023-translation} and accompanied with preceding context. We provide these to native or bilingual speakers \textemdash who voluntarily participated in the annotation process\textemdash{} as we are interested in how non-expert translators from the general public perceive the translations. We mask the MT type information~\footnote{Annotators see the translation hypotheses as pairs of \texttt{output\_1, output\_2}} and ask them to annotate the data as follows: first, identify any of the four linguistic phenomena present in the source sentence. Once identified, they determine whether the phenomenon is translated correctly in (a) the greedy translation and (b) the QAD translation. Next, they annotate the semantic difference between the greedy and QAD hypotheses using a Likert scale ranging from 1 to 5. They then select their preferred translation between the two hypotheses and may optionally provide comments explaining their preference and observations (full guidelines in \Cref{human-details-appendix}).

 Annotation statistics are presented in \Cref{dataset_statistics_appendix}, where we see a high correlation between automatic tags with MuDA and human tags (with an overlap of 60\%-100\%).
\Cref{performance-plot} presents the average performance of greedy and QAD outputs across languages, showing improved performance for QAD across all phenomena, which aligns with the results of the automatic evaluation. 

Results of the semantic difference against preferences are presented in \Cref{preference-plot} (Arabic is excluded from these figures to remove model-specific bias; its results in \Cref{arabic-appendix} confirm the same findings). We notice that 
QAD output is generally preferred, while greedy output tends to be less frequently chosen as the preferred option, especially when there are larger semantic differences between the outputs. Greedy output is still sometimes preferred in cases where the semantic differences are smaller. These patterns suggest that \textbf{QAD generates semantically richer samples that align with human preferences compared to greedy decoding}. In addition, analyzing the comments we received from the participants, it seems that QAD-based outputs are closer to human perception in terms of discourse and fluency, even when translation errors occur.

Finally, in \Cref{qualitative_examples}, we present the details and findings of a qualitative analysis on TED2020 data. We analyze sentence-level and context-aware greedy and QAD outputs and showcase specific examples demonstrating how the context-aware QAD setup unlocks models' ability to handle different discourse phenomena such as pronoun disambiguation, lexical repetition, formality and verb forms.

\section{Related Work}
\label{related-work}
\paragraph{Document-level translation.} Document-level translation models allow incorporating inter-sentential dependencies during translation. Architectures include single-encoder approaches \citep{tiedemann-scherrer-2017-neural,bawden-etal-2018-evaluating}, multi-encoder approaches \citep{libovicky-helcl-2017-attention,zoph-knight-2016-multi,wang-etal-2017-exploiting-cross,bawden-etal-2018-evaluating,zhang-etal-2018-improving,li-etal-2020-multi-encoder,lupo-etal-2022-divide}, multi-hop transformers \citep{zhang-etal-2021-multi},  hierarchical context-encoders \citep{wang-etal-2017-exploiting-cross,miculicich-etal-2018-document,maruf-etal-2019-selective,DBLP:conf/aaai/YunHJ20}, 
and document-embedding approaches \citep{huo-etal-2020-diving,mace-servan-2019-using,morishita-etal-2021-context}.

\paragraph{LLMs for document-level translation.} 
LLMs are becoming increasingly competitive in MT \citep{kocmi-etal-2024-findings}. Efforts to adapt LLMs for document-level MT include finetuning models using mixed sentence-level and document-level instructions \citep{DBLP:journals/corr/abs-2401-08088,DBLP:journals/corr/abs-2504-12140}, prompting models via in-context learning \citep{cui-etal-2024-efficiently}, hybrid techniques combining sentence-level translation models and monolingual document-level language models \citep{petrick-etal-2023-document}, and agentic systems based on multi-level memory \citep{DBLP:conf/iclr/WangZLWMZZ25}.

\paragraph{Discourse evaluation in machine translation.} 
Studying discourse characteristics in MT has been going on since the early works on rule-based and statistical MT systems \citep{hardmeier2012discourse} and continued as NMT models prevailed \citep{tan2022towards,honda2023context,luo2024context}. Subsequent studies have developed discourse-specific datasets and benchmarks, including test sets for coreference, entities, terminology, quotations, and readability \citep{jiang-etal-2023-discourse,DBLP:journals/corr/abs-2004-14607,muller-etal-2018-large,lopes-etal-2020-document}. Research efforts have also focused on designing evaluation metrics to assess discourse performance, such as cohesion, coreference resolution, terminology consistency, and zero pronoun translation \citep{DBLP:journals/corr/abs-2208-09118,bawden-etal-2018-evaluating,DBLP:journals/corr/abs-2401-06468,wang-etal-2023-document-level}.\looseness-1

\section{Conclusion}
We investigate the discourse phenomena performance of LLMs in context-aware translation. Specifically, we examine four aspects of discourse: lexical repetition, formality, pronoun resolution, and verb forms. Our findings reveal that LLMs still exhibit weaknesses in discourse performance when using greedy decoding. To address this limitation, we propose the use of quality-aware decoding (QAD) to better leverage the discourse knowledge encoded within LLMs. We demonstrate the effectiveness of QAD through extensive automatic evaluations across six language pairs and two datasets. Additionally, we conduct an ablation study comparing different decoding methods and perform a human assessment on a subset of the data to analyze the lexical and semantic changes introduced by QAD. To support further research, we release the dataset with human annotations of discourse phenomena. Future research can explore the use of such annotated data as a reward signal for fine-tuning LLMs to further enhance their discourse phenomena performance.

\section*{Limitations}
\begin{itemize}

    \item We use MuDA \citep{fernandes-etal-2023-translation} as it provides an automatic method for tagging various discourse phenomena. However, it does not capture all aspects of discourse in translation such as zero pronoun translation, readability, etc. Additionally, we adopt MuDA's default alignment and coreference resolution models, which may not reflect the current state of the art. Enhancing these components and employing a more comprehensive discourse evaluation are directions for future research.

    \item We use only one baseline decoding method which is greedy decoding. We leave it to future work to experiment on other baselines such as beam search, as prior studies have shown its effectiveness, particularly for encoder–decoder models \citep{DBLP:conf/nips/SutskeverVL14,freitag-al-onaizan-2017-beam}.
   
    \item We experiment with only one sampling approach and one hyperparameter setup for it (nucleus sampling with p=0.9). We leave it to future research to investigate the effect of the number of samples, the sampling method used and its hyper-parameters on discourse performance.

    \item We perform the LLM-as-a-judge evaluation at the overall translation level, as we utilize an off-the-shelf model that was not sensitive to specific phenomena changes. Future work could focus on adapting LLM judges to discourse phenomena evaluation.

    \item We attempt to cover as many languages and models as possible, given the experimental resources we have. Additional observations may arise for languages and models we did not cover.

    \item We perform the human evaluation on a limited amount of data. Based on our conclusions, it would be useful to have a larger dataset with human annotations, which would allow for more comprehensive experiments, supervision of models, etc.

    \item The performance on the discourse phenomena we study can be affected by gender bias in the tested models, we leave it to future work to study the correlation between models' bias and discourse performance.

\end{itemize}

\section*{Ethical Considerations}
Machine translation is a widely adopted technology, sometimes in sensitive, high-risk settings. Even though we perform a thorough analysis of LLMs' performance on discourse phenomena during translation, and propose the use of quality aware-decoding to improve the performance, we still rely heavily on automatic evaluation which is imperfect. For systems deployed in critical scenarios, we advocate for detailed, case-specific assessments to ensure reliability.

\section*{Acknowledgements}
We would like to thank Hippolyte Gisserot-Boukhlef, 
Sergey Troshin, Beatriz Canaverde and Emmanouil Zaranis for their valuable discussions about this work, and their help with the annotations. This work is part of the UTTER project, supported by the European Union's Horizon Europe research and innovation programme via grant agreement 101070631.
VN acknowledges support from the Dutch Research Council (NWO) via VI.Veni.212.228. 
WM acknowledges travel support from ELIAS (GA no 101120237).
CZ was partially funded by the FCT project ``OptiGov'' (DOI 10.54499/2024.07385.IACDC), funded by the PRR under the measure RE-C05-i08.m04. She is also supported by the PRR Plan through projects C645008882-00000055 (Center for Responsible AI) and UID/50008: Instituto de Telecomunicações. Finally, she supported by an unrestricted gift from Google (Google Research Scholar).

\bibliography{anthology,custom}

\appendix

\section{Dataset statistics}
\label{dataset_statistics_appendix}
\Cref{dataset_statistics} presents the dataset statistics for the three corpora (DELA, TED2020, WMT24++), including discourse phenomena statistics as well as the number of sentences and documents. \Cref{human_phenomena_statistics} presents the annotation statistics of discourse phenomena in the human assessment analysis.

\section{WMT24++ Results}
\label{wmt24_results_appendix}
\begin{figure}
\centering
\includegraphics[width=\columnwidth]{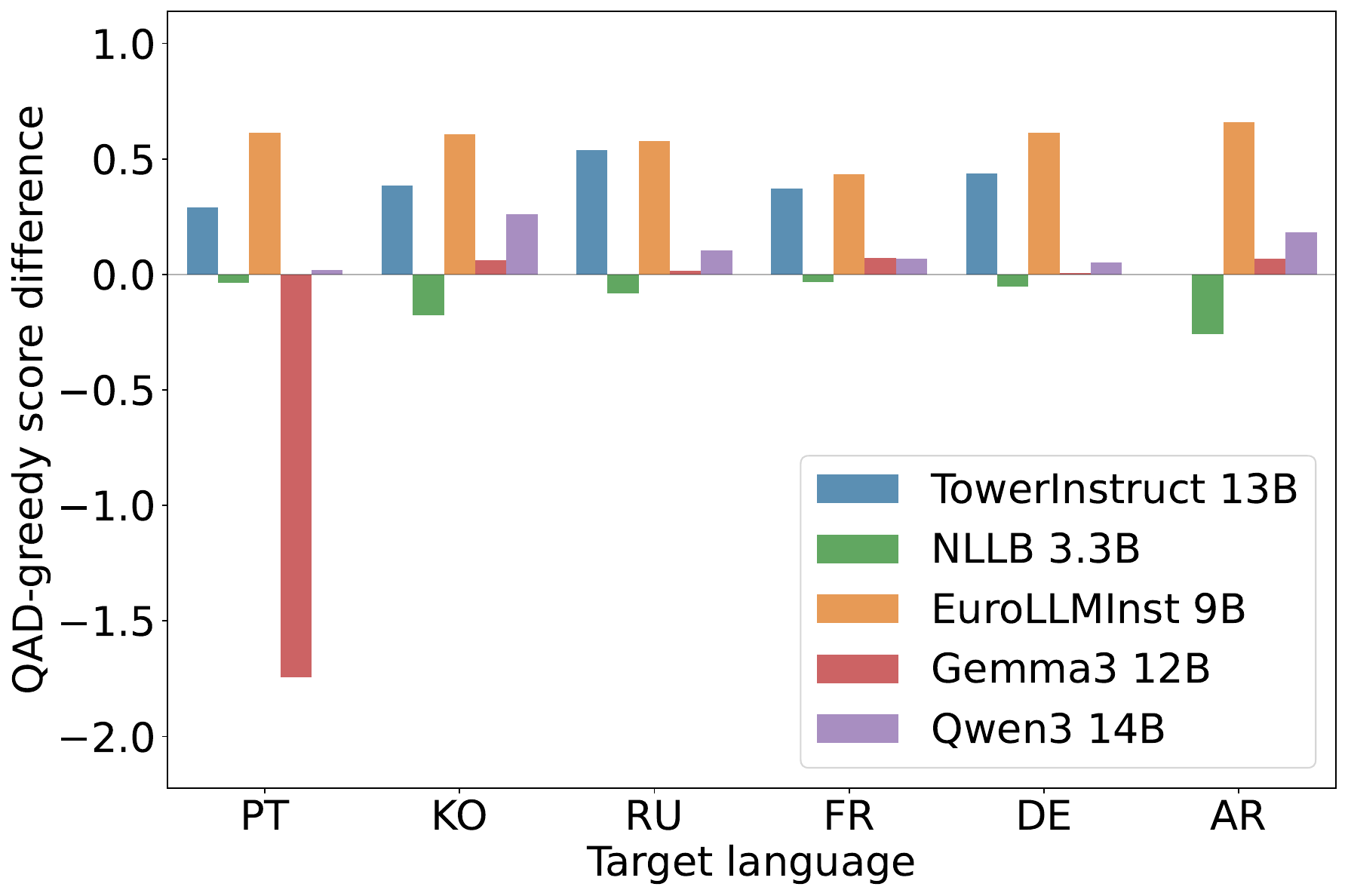}
\caption{Difference between QAD and greedy LLM-as-a-judge scores on WMT24++ data. The plot demonstrates that QAD improves the performance of LLMs.} 
\label{llm-as-a-judge-plot-wmt}
\end{figure}

\Cref{averaged_wmt_results} shows the results on WMT24 ++ for averaged across all language pairs. LLM-as-a-judge scores are shown in \Cref{llm-as-a-judge-plot-wmt}. 

\section{Detailed Language-Specific Results}
\label{appendix-language-specific}
Detailed language-specific results for all models on TED2020 and WMT24++ datasets are shown in \Cref{ted_results,wmt_results}, respectively. 
It is worth mentioning that all models exhibit low performance on {\enge} WMT24++ data. A manual qualitative analysis of the translations reveals that the reference translations are of suboptimal quality, often consisting of short sentences.

\section{Results of Human Qualitative Analysis on Arabic}
\label{arabic-appendix}
\begin{figure}
\centering
\includegraphics[width=\columnwidth]{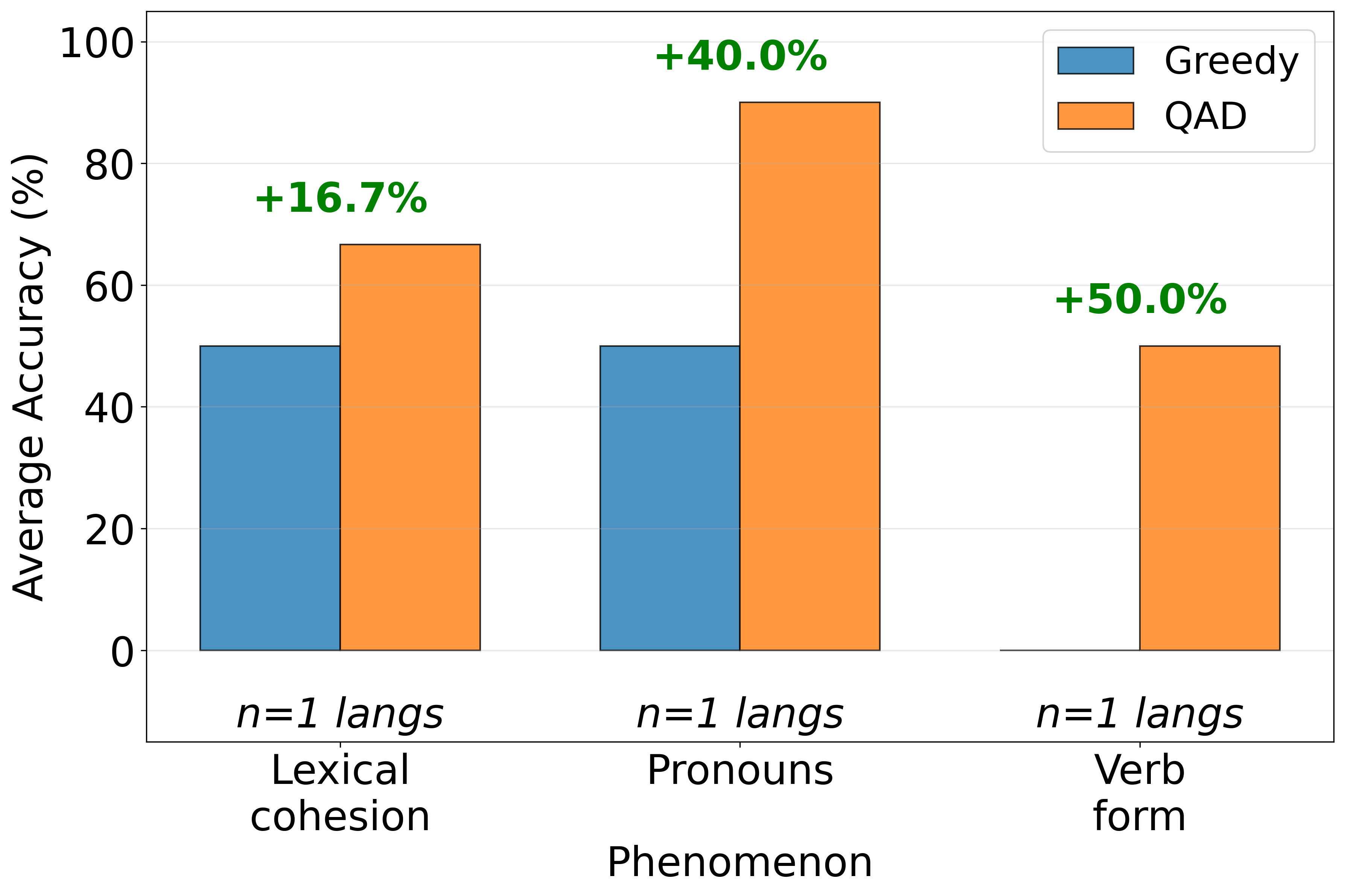}
\caption{Human-annotated accuracy of greedy and QAD outputs in handling discourse phenomena for Arabic data.}
\label{performance-plot-arabic}
\end{figure}

\begin{figure}
\centering
\includegraphics[width=\columnwidth]{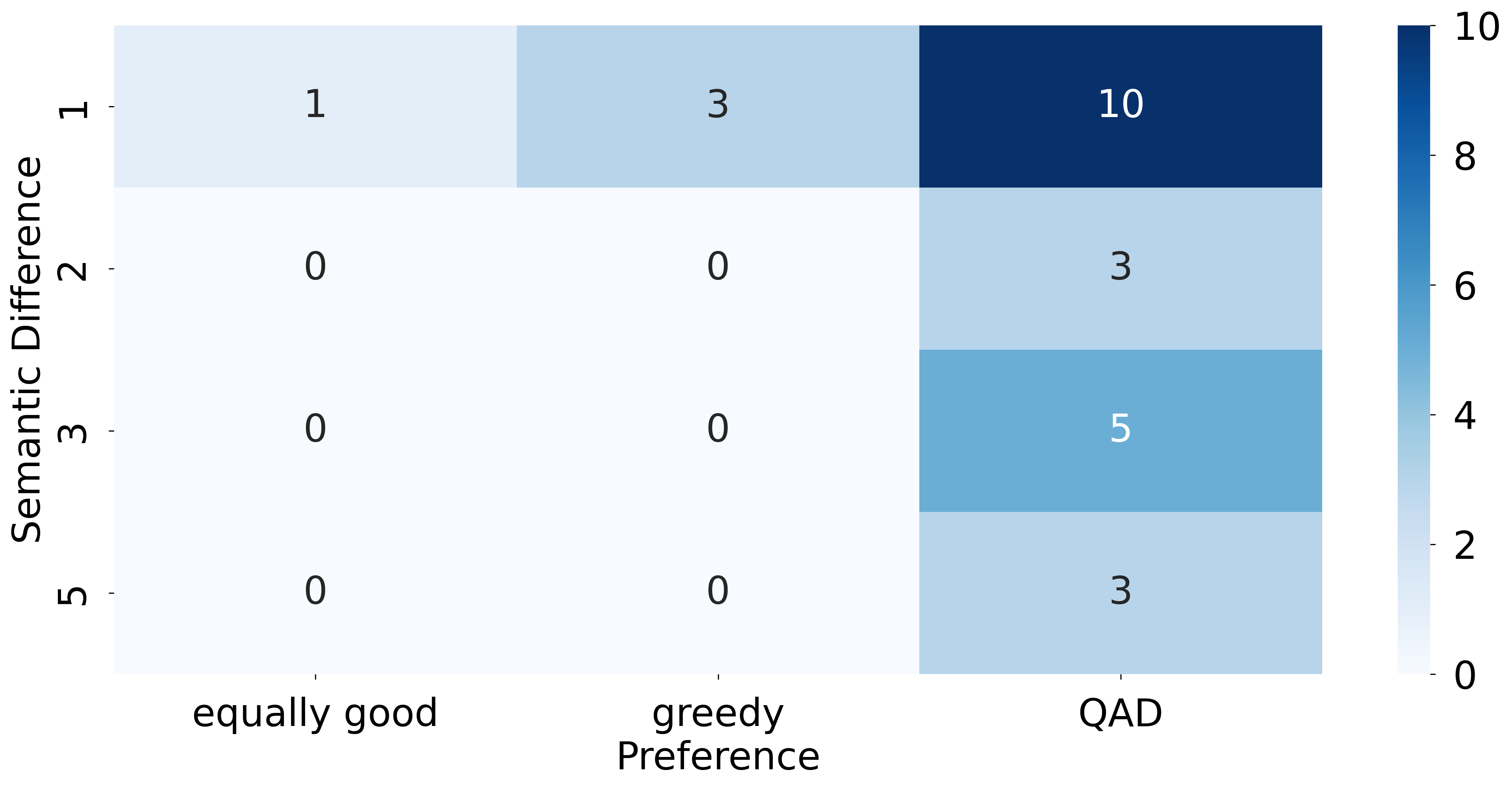}
\caption{Semantic difference vs. preference on Arabic data.} \label{preference-plot-arabic}
\end{figure}

\Cref{performance-plot-arabic} shows the human annotations of the performance of QAD and greedy outputs on Arabic. \Cref{preference-plot-arabic} shows the preference and semantic difference relationship for Arabic.

\section{Prompt Formats}
\label{prompt-formats-appendix}
\Cref{tower-prompt-format,eurollm-prompt-format,gemma-qwen-prompt-format} present the prompt formats used to prompt the models.

\section{Human Assessment Details}
\label{human-details-appendix}
Details of the data and instructions given to the annotators are presented in \Cref{Human_assessment_details}.

\section{Sustainability Statement}
\label{sustainability}
Our experiments run in 782h on 1 GPU NVIDIA A100 40GB PCIe, and draw 334.06 kWh. 
Based in [redacted for anonymity],
this has a carbon footprint of 125.05 kg CO2e, which is equivalent to 11.37 tree-years \citep{lannelongue2021green}.

\section{SacreBLEU Signatures}
\label{metric-signatures}
To ensure reproducability, we present the SacreBLEU signatures for BLEU and ChrF metrics in \Cref{signatures}.

\begin{table*}
    \centering\small
    \begin{tabular}{ll}
    \toprule
        metric & signature \\
        \midrule
        \BLEU & nrefs:1|case:mixed|eff:no|tok:13a|smooth:exp|version:2.5.1 \\
        \CHRF & nrefs:1|case:mixed|eff:yes|nc:6|nw:0|space:no|version:2.5.1 \\
        \bottomrule
    \end{tabular}
    \caption{Evaluation-metrics signatures}
    \label{signatures}
\end{table*}

\begin{table*}
\centering
\small
\begin{tabular}{ l r r r r r r r}
\toprule
 & Lexical repetition & Formality & Pronouns & Verb form & Total & Sentences & Documents \\
\midrule
\multicolumn{8}{l}{\textbf{DELA}}\\ 
 \enpt & 1322 & 630 & 323 & -- & 1866 (50.3) & 3710 & 60 \\
\midrule
\multicolumn{8}{l}{\textbf{TED2020}}\\
 \enpt & 6640 & 3151 & 2202 & -- & 9877 (49.4) & 20003 & 162 \\
 \enge & 5386 & 4904 & 2186 & -- & 10125 (50.4) & 20077 & 160 \\
 \enfr & 6346 & 3315 & 7486 & -- & 11642 (58.1) & 20049 & 162 \\
 \enko & 2190 & 1165 & -- & -- & 3238 (16.2) & 20017 & 162 \\
 \enar & 4109 & -- & 655 & -- & 4654 (23.2) & 20034 & 162 \\
 \enru & 3544 & 2451 & -- & -- & 5506 (27.4) & 20084 & 163 \\
\midrule
\multicolumn{8}{l}{\textbf{WMT24}}\\
 \enpt & 209 & 178 & 59 & -- & 356 (37.1) & 960 & 169 \\
 \enge & 56 & 199 & 43 & -- & 263 (27.4) & 960 & 169 \\
 \enfr & 189 & 130 & 160 & 67 & 413 (43.0) & 960 & 169 \\
 \enko & 93 & 17 & -- & -- & 109 (11.4) & 960 & 169 \\
 \enar & 166 & -- & 39 & -- & 198 (20.6) & 960 & 169 \\
 \enru & 129 & 90 & -- & 70 & 255 (26.6) & 960 & 169 \\
\bottomrule
\end{tabular}
\caption{Dataset statistics, including counts of each phenomenon, the total number of sentences tagged with phenomena and their percentage of total sentences (in parentheses), and the total number of sentences and documents for each dataset and language pair. Note that the Total column can be less than the sum of phenomena columns because we can have multiple phenomena per sentence.}
\label{dataset_statistics}
\end{table*}
\begin{table}
\centering
\small
\begin{tabular}{ l r r r r r r}
\toprule
 & rep. & Formal. & Pro. & Verb. & Total & (\%)\\
\midrule
\enpt & 12  & 1  & 7  & 1  & 15 & 60 \\
\enge & 16  & 5  & 15  & 6  & 22 & 88 \\
\enfr & 5  & 14  & 10  & 9  & 21 & 84 \\
\enko & 0  & 2  & 4  & 25  & 25 & 100 \\
\enar & 6  & 0  & 10  & 2  & 16 & 64 \\
\enru & 4  & 7  & 7  & 6  & 16 & 64 \\

\bottomrule
\end{tabular}
\caption{Human-annotated discourse phenomena statistics, including counts of each phenomenon, the total number of sentences tagged with phenomena and their percentage of total sentences. Note that the total column can be less than the sum of phenomena columns because we can have multiple phenomena per sentence.
}
\label{human_phenomena_statistics}
\end{table}
\begin{table*}
\centering
\small
\setlength{\tabcolsep}{4pt}
\begin{tabular}{lccccccccc}
\toprule
& \BLEU & \COMET & \docCOMET & \COMETQE & \docCOMETQE & L.repetition & Formality & Pronouns & Verb form \\
\midrule
\textbf{ctx= 0} \\
\textcolor{gray}{nllb G} & \textcolor{gray}{20.3} & \textcolor{gray}{72.5} & \textcolor{gray}{70.3} & \textcolor{gray}{75.8} & \textcolor{gray}{67.4} & \textcolor{gray}{56.5} & \textcolor{gray}{40.8} & \textcolor{gray}{44.2} & \textcolor{gray}{34.5} \\
\textcolor{gray}{nllb Q} & \textcolor{gray}{22.2} & \textcolor{gray}{73.3} & \textcolor{gray}{71.0} & \textcolor{gray}{76.5} & \textcolor{gray}{68.6} & \textcolor{gray}{53.3} \textbf{$\color{red}\downarrow$} & \textcolor{gray}{43.6} & \textcolor{gray}{42.2} \textbf{$\color{red}\downarrow$} & \textcolor{gray}{41.5} \\
tower G & 16.9 & 72.3 & 69.9 & {74.8} & {66.3} & 49.8 & 33.8 & 28.7 & {26.5} \\
tower Q & 24.8 & 76.2 & 73.8 & {\textbf{80.6}} & {\textbf{73.8}} & 56.6 & 43.4 & 36.7 & 39.5 \\
euro G & 14.1 & 67.9 & 65.1 & {72.9} & 63.0 & 45.7 & 36.4 & 36.8 & 28.5 \\
euro Q & 21.5 & 72.0 & 69.2 & {79.6} & {72.6} & 57.5 & 45.4 & 41.2 & \textbf{43.0} \\
gemma G & 21.2 & {71.7} & {68.9} & {78.0}  & {71.4}  & 52.8  & {44.2} & 41.0 & {41.0}  \\
gemma Q & 22.3 & {73.0} & {70.2} & {79.7} & {72.9} & 55.3 & 48.4 & {46.0} & {\textbf{43.0}} \\
qwen G & 19.5 & 71.4 & 68.8 & 78.2 & 70.8 &  52.3 & 41.2 & 44.0 & 36.5 \\
qwen Q & 20.8 & 72.3 & 69.6 & {79.8} & 72.9 & 54.7 & 40.0 \textbf{$\color{red}\downarrow$} & 43.2 \textbf{$\color{red}\downarrow$} & {39.5} \\
\midrule
\textbf{ctx= 5} \\
tower G & \newuline{17.0} & \newuline{73.0} & \newuline{71.5} & 69.0 & 63.5 & \newuline{54.2} & \newuline{39.8} & \newuline{38.0} & 25.5 \\
tower Q & \newuline{\textbf{25.7}} & \newuline{\textbf{78.5}} & \newuline{\textbf{76.7}} & 76.0 & 72.0 & \newuline{60.2} & \newuline{\textbf{49.2}} & \newuline{41.0} & \newuline{41.5} \\
euro G & \newuline{15.9} & \newuline{71.9} & \newuline{69.6} & 71.6 & \newuline{63.2} & \newuline{55.8} & \newuline{37.8} & \newuline{38.8} & \newuline{32.5} \\
euro Q & \newuline{24.9} & \newuline{77.1} & \newuline{74.8} & 78.5 & 72.1 & \newuline{\textbf{65.3}} & \newuline{46.6} & \newuline{46.5} & \newuline{\textbf{43.0}} \\
gemma G & \newuline{22.1} & 71.0 & 68.4 & 75.7 & 70.4 & \newuline{61.5} & 43.6 & \newuline{42.2} & 38.5 \\
gemma Q & \newuline{23.4} & 72.6 & 70.1 & 78.5 & 72.5 & \newuline{64.7} & \newuline{50.8} & 43.0 & 42.5 \\
qwen G & \newuline{20.5} & \newuline{72.1} & \newuline{69.6} & \newuline{78.2} & \newuline{71.1} & \newuline{62.5} & \newuline{47.6} & \newuline{44.8} & \newuline{38.5} \\
qwen Q & \newuline{21.7} & \newuline{72.8} & \newuline{70.4} & 79.6 & \newuline{73.0} & \newuline{64.2} & \newuline{42.2} \textbf{$\color{red}\downarrow$} & \newuline{\textbf{48.0}} & 39.0 \\
\bottomrule
\end{tabular}
\caption{Translation and discourse phenomena performance of all models (nllb={\nllb}, tower={\towerinst}, euro={\eurollm}, gemma={\gemmathree}, qwen={\qwenthree}) using greedy (G) and quality-aware decoding (Q) on \textbf{WMT24++ dataset} in both sentence-level (ctx= 0) and context-aware (ctx= 5) setups. The results are averaged across all language pairs. \textbf{Bold} highlights the best value per column. Cases where the QAD result is worse than its Greedy counterpart are marked with a red down arrow \textbf{$\color{red}{\downarrow}$}. Cases where the context-aware result is better than the sentence level result are marked with a \newuline{green underline}.}
\label{averaged_wmt_results}
\end{table*}
\begin{table*}
\centering
\tiny
\setlength{\tabcolsep}{4pt}
\begin{tabular}{llcccccccccccccccccc}
\toprule
& \multirow{3}{*}{\textbf{Metric}} & \multicolumn{2}{c}{\textbf{nllb}} & \multicolumn{4}{c}{\textbf{tower}} & \multicolumn{4}{c}{\textbf{euro}}  & \multicolumn{4}{c}{\textbf{gemma}}  & \multicolumn{4}{c}{\textbf{qwen}} \\
  & & \multicolumn{2}{c}{\textbf{ctx= 0}} & \multicolumn{2}{c}{\textbf{ctx= 0}} & \multicolumn{2}{c}{\textbf{ctx= 5}} & \multicolumn{2}{c}{\textbf{ctx= 0}} & \multicolumn{2}{c}{\textbf{ctx= 5}} & \multicolumn{2}{c}{\textbf{ctx= 0}} & \multicolumn{2}{c}{\textbf{ctx= 5}} & \multicolumn{2}{c}{\textbf{ctx= 0}} & \multicolumn{2}{c}{\textbf{ctx= 5}} \\
& & \textbf{G} & \textbf{Q} & \textbf{G} & \textbf{Q} & \textbf{G} & \textbf{Q} & \textbf{G} & \textbf{Q} & \textbf{G} & \textbf{Q} & \textbf{G} & \textbf{Q} & \textbf{G} & \textbf{Q} & \textbf{G} & \textbf{Q} & \textbf{G} & \textbf{Q} \\
\midrule
\multirow{8}{*}{\textbf{EN-PT}} & \BLEU & 40.4 & 41.8 & 26.5 & 38.5 & 30.4 & \textbf{42.5} & 24.7 & 37.4 & 21.0 & 38.9 & 38.1 & 39.2 & 40.2 & 39.1 & 38.3 & 39.2 & 40.1 & 41.0 \\
& \COMET & 87.0 & 87.2 & 83.6 & 87.5 & 84.6 & \textbf{88.2} & 82.7 & 86.8 & 81.1 & 87.2 & 85.0 & 87.1 & 85.8 & 87.1 & 87.1 & 87.3 & 87.6 & 87.8 \\
& \docCOMET & 82.3 & 82.5 & 78.0 & 82.6 & 79.8 & \textbf{83.9} & 75.6 & 80.4 & 74.2 & 81.1 & 79.7 & 82.2 & 80.7 & 82.2 & 82.2 & 82.5 & 82.9 & 83.2 \\
& \COMETQE & 82.7 & 82.9 & 80.1 & \textbf{83.6} & 78.8 & 82.2 & 79.6 & 83.5 & 77.7 & 83.3 & 80.8 & 83.2 & 81.4 & 83.2 & 83.0 & 83.2 & 82.9 & 83.1 \\
& \docCOMETQE & 80.1 & 80.5 & 75.5 & \textbf{81.4} & 74.1 & 79.7 & 73.2 & 81.3 & 68.9 & 81.1 & 78.5 & 80.8 & 78.7 & 80.8 & 80.6 & 81.0 & 80.7 & 81.1 \\
& L.repetition & 80.0 & 80.0 & 73.0 & 79.0 & 78.0 & \textbf{83.0} & 73.0 & 80.0 & 75.0 & \textbf{83.0} & 76.0 & 78.0 & 80.0 & 78.0 & 79.0 & 80.0 & 82.0 & 82.0 \\
& Formality & 65.0 & 67.0 & 46.0 & 62.0 & 58.0 & \textbf{69.0} & 56.0 & 66.0 & 53.0 & 68.0 & 19.0 & 64.0 & 38.0 & 64.0 & 64.0 & 65.0 & 66.0 & 67.0 \\
& Pronouns & 51.0 & 51.0 & 34.0 & 43.0 & 51.0 & \textbf{61.0} & 41.0 & 49.0 & 47.0 & 57.0 & 45.0 & 47.0 & 54.0 & 47.0 & 48.0 & 48.0 & 55.0 & 55.0 \\
\midrule
\multirow{8}{*}{\textbf{\enge}} & \BLEU & 31.3 & 32.7 & 20.9 & 31.7 & 21.9 & \textbf{33.1} & 15.6 & 29.0 & 14.1 & 29.2 & 28.1 & 29.7 & 29.0 & 31.8 & 27.3 & 28.3 & 29.4 & 30.4 \\
& \COMET & 83.8 & 84.1 & 79.7 & 84.6 & 79.9 & \textbf{85.1} & 77.4 & 83.9 & 76.2 & 84.1 & 79.8 & 83.8 & 77.4 & 82.4 & 83.3 & 83.8 & 84.0 & 84.5 \\
& \docCOMET & 79.1 & 79.4 & 74.3 & 79.9 & 75.3 & \textbf{80.9} & 70.5 & 78.0 & 70.2 & 79.1 & 74.3 & 78.8 & 71.9 & 77.4 & 78.4 & 78.9 & 79.4 & 79.9 \\
& \COMETQE & 82.9 & 82.9 & 78.8 & 83.3 & 76.8 & 81.9 & 77.6 & \textbf{83.4} & 76.5 & 83.0 & 78.3 & 82.7 & 75.6 & 80.6 & 82.7 & 83.2 & 82.7 & 83.1 \\
& \docCOMETQE & 80.8 & 81.0 & 74.3 & \textbf{81.6} & 72.3 & 80.1 & 68.9 & 81.2 & 67.9 & 80.8 & 76.5 & 80.7 & 73.1 & 78.9 & 80.7 & 81.4 & 80.9 & 81.5 \\
& L. repetition & 69.0 & 69.0 & 63.0 & 70.0 & 68.0 & \textbf{76.0} & 60.0 & 68.8 & 64.0 & 72.0 & 65.0 & 67.0 & 66.0 & 71.0 & 66.0 & 67.0 & 71.0 & 71.0 \\
& Formality & 65.0 & 67.0 & 62.0 & 70.0 & 67.0 & \textbf{75.0} & 56.0 & 68.0 & 58.0 & 70.0 & 10.0 & 60.0 & 09.0 & 30.0 & 56.0 & 57.0 & 70.0 & 71.0 \\
& Pronouns & 68.0 & 67.0 & 56.0 & 65.0 & 63.0 & \textbf{73.0} & 55.0 & 66.0 & 59.0 & 69.0 & 63.0 & 66.0 & 62.0 & 66.0 & 66.0 & 66.0 & 67.0 & 68.0 \\
\midrule
\multirow{9}{*}{\textbf{\enfr}} & \BLEU & 41.0 & \textbf{43.0} & 27.2 & 40.3 & 31.1 & 42.9 & 21.2 & 37.2 & 20.5 & 38.7 & 36.8 & 38.3 & 37.7 & 39.1 & 37.2 & 38.3 & 39.2 & 40.1 \\
& \COMET & 84.0 & 84.5 & 80.9 & 85.1 & 81.6 & \textbf{85.7} & 77.7 & 84.3 & 76.8 & 84.5 & 81.4 & 84.3 & 78.6 & 82.3 & 84.1 & 84.3 & 84.7 & 85.0 \\
& \docCOMET & 80.0 & 80.6 & 76.4 & 81.2 & 77.8 & \textbf{82.3} & 71.6 & 79.3 & 71.4 & 80.1 & 76.9 & 80.1 & 73.9 & 78.0 & 80.1 & 80.1 & 80.9 & 81.3 \\
& \COMETQE & 84.1 & 84.4 & 82.2 & \textbf{85.3} & 80.9 & 84.1 & 79.6 & 84.9 & 78.1 & 84.6 & 81.7 & 84.7 & 78.9 & 83.0 & 84.7 & 84.7 & 84.7 & 84.9 \\
& \docCOMETQE & 81.7 & 82.5 & 78.1 & \textbf{83.4} & 77.1 & 82.2 & 71.1 & 82.5 & 68.9 & 82.2 & 80.1 & 82.4 & 76.6 & 80.6 & 82.6 & 82.4 & 82.8 & 83.2 \\
& L. repetition & 78.0 & 79.0 & 71.0 & 77.0 & 76.0 & \textbf{81.0} & 69.0 & 77.0 & 70.0 & 79.0 & 74.0 & 77.0 & 72.0 & 77.0 & 77.0 & 77.0 & 79.0 & 79.0 \\
& Formality & 75.0 & 74.0 & 66.0 & 75.0 & 71.0 & \textbf{79.0} & 60.0 & 75.0 & 61.0 & 76.0 & 18.0 & 71.0 & 11.0 & 51.0 & 66.0 & 66.0 & 77.0 & 77.0 \\
& Pronouns & 75.0 & 75.0 & 64.0 & 73.0 & 72.0 & \textbf{79.0} & 63.0 & 73.0 & 64.0 & 76.0 & 66.0 & 69.0 & 69.0 & 72.0 & 71.0 & 72.0 & 77.0 & 78.0 \\
\midrule
\multirow{7}{*}{\textbf{\enru}} & \BLEU & 24.2 & 24.9 & 16.1 & 24.4 & 11.7 & \textbf{26.2} & 14.7 & 23.8 & 15.6 & 25.4 & 21.5 & 22.4 & 23.0 & 24.2 & 21.1 & 21.8 & 22.7 & 23.7 \\
& \COMET & 84.3 & 84.3 & 80.6 & 85.1 & 71.6 & \textbf{85.8} & 81.1 & 85.2 & 81.0 & 85.7 & 83.3 & 84.5 & 82.6 & 84.1 & 83.8 & 84.2 & 84.8 & 85.2 \\
& \docCOMET & 79.7 & 79.8 & 75.5 & 80.5 & 66.9 & \textbf{81.8} & 75.8 & 80.6 & 76.0 & 81.2 & 78.2 & 79.6 & 77.8 & 79.5 & 79.1 & 79.5 & 80.4 & 80.8 \\
& \COMETQE & 82.7 & 82.6 & 78.1 & 83.3 & 64.1 & 81.8 & 78.9 & \textbf{83.5} & 78.8 & 83.3 & 82.2 & 83.2 & 80.9 & 82.4 & 82.6 & 83.1 & 82.7 & 83.1 \\
& \docCOMETQE & 80.0 & 79.8 & 71.7 & 81.0 & 65.6 & 79.4 & 72.2 & 81.2 & 72.7 & \textbf{81.3} & 79.6 & 80.7 & 78.3 & 80.1 & 79.9 & 80.7 & 80.3 & 81.1 \\
& L. repetition & 58.0 & 59.0 & 53.0 & 59.0 & 44.0 & \textbf{64.0} & 52.0 & 58.0 & 56.0 & 62.0 & 55.0 & 56.0 & 58.0 & 60.0 & 57.0 & 62.0 & 62.0 & 62.0 \\
& Formality & 56.0 & 56.0 & 47.0 & 57.0 & 39.0 & \textbf{61.0} & 48.0 & 58.0 & 48.0 & 60.0 & 31.0 & 50.0 & 21.0 & 42.0 & 51.0 & 60.0 & 59.0 & 60.0 \\
\midrule
\multirow{7}{*}{\textbf{\enar}} & \BLEU & 12.5 & 12.5 & N/A & N/A & N/A & N/A & 6.5 & 11.4 & 5.2 & \textbf{13.4} & 09.3 & 10.2 & 11.0 & 10.2 & 08.2 & 09.6 & 09.5 & 10.9 \\
& \COMET & 81.3 & 81.2 & N/A & N/A & N/A & N/A & 77.2 & 82.1 & 75.0 & \textbf{82.5} & 80.1 & 81.5 & 78.0 & 81.6 & 77.9 & 80.9 & 79.3 & 81.8 \\
& \docCOMET & 75.0 & 74.9 & N/A & N/A & N/A & N/A & 69.9 & 75.5 & 68.3 & \textbf{76.2} & 72.9 & 74.6 & 71.1 & 74.7 & 70.8 & 74.0 & 72.6 & 75.4 \\
& \COMETQE & 79.1 & 78.7 & N/A & N/A & N/A & N/A & 74.0 & \textbf{80.3} & 70.8 & 79.5 & 78.9 & \textbf{80.3} & 75.8 & \textbf{80.3} & 75.9 & 79.3 & 76.3 & 79.3 \\
& \docCOMETQE & 76.6 & 75.9 & N/A & N/A & N/A & N/A & 68.0 & \textbf{78.4} & 61.8 & 77.4 & 76.5 & 78.3 & 73.5 & 78.2 & 70.5 & 76.8 & 71.7 & 77.1 \\
& L. repetition & 55.0 & 55.0 & N/A & N/A & N/A & N/A & 48.0 & 54.0 & 53.0 & \textbf{60.0} & 52.0 & 53.0 & 54.0 & 53.0 & 50.0 & 51.0 & 57.0 & 58.0 \\
& Pronouns & \textbf{51.0} & 49.0 & N/A & N/A & N/A & N/A & 41.0 & 48.0 & 41.0 & 50.0 & 47.0 & 48.0 & 50.0 & 48.0 & 46.0 & 50.0 & 46.0 & 50.0 \\
\midrule
\multirow{7}{*}{\textbf{\enko}} & \BLEU & 20.6 & 20.9 & 14.2 & 20.8 & 9.7 & 20.3 & 12.1 & 20.9 & 13.6 & \textbf{23.7} & 16.9 & 18.7 & 20.3 & 22.4 & 16.4 & 18.3 & 19.4 & 21.4 \\
& \COMET & 84.7 & 84.7 & 82.9 & 86.1 & 80.1 & 85.9 & 81.3 & 85.7 & 82.0 & 86.8 & 81.8 & 85.3 & 83.0 & 85.0 & 84.5 & 85.8 & 85.8 & \textbf{86.9} \\
& \docCOMET & 79.1 & 79.0 & 76.4 & 80.4 & 74.4 & 80.9 & 74.8 & 80.1 & 76.0 & 81.6 & 75.2 & 79.1 & 77.1 & 79.5 & 78.2 & 79.7 & 80.4 & \textbf{81.7} \\
& \COMETQE & 84.7 & 84.4 & 81.4 & 85.6 & 74.7 & 82.6 & 80.3 & 85.5 & 79.9 & 85.4 & 81.3 & 85.0 & 82.1 & 84.4 & 84.3 & \textbf{85.7} & 84.5 & \textbf{85.7} \\
& \docCOMETQE & 80.9 & 80.4 & 75.8 & 82.3 & 67.7 & 79.2 & 72.7 & 82.1 & 73.4 & 82.2 & 78.6 & 81.5 & 79.2 & 81.3 & 80.0 & 82.3 & 80.7 & \textbf{82.6} \\
& L. repetition & 45.0 & 46.0 & 40.0 & 47.0 & 44.0 & \textbf{52.0} & 39.0 & 45.0 & 45.0 & 50.0 & 40.0 & 44.0 & 49.0 & 50.0 & 42.0 & 43.0 & 49.0 & 50.0 \\
& Formality & 26.0 & 24.0 & 23.0 & 27.0 & 26.0 & \textbf{39.0} & 20.0 & 28.0 & 27.0 & 38.0 & 04.0 & 32.0 & 13.0 & 26.0 & 15.0 & 13.0 & 33.0 & 35.0 \\
\bottomrule
\end{tabular}
\caption{Detailed language-specific translation and discourse phenomena performance of all models (nllb={\nllb}, tower={\towerinst}, euro={\eurollm}, gemma={\gemmathree}, qwen={\qwenthree}) using greedy (G) and quality-aware decoding (Q) on \textbf{TED2020 dataset} in both sentence-level (ctx= 0) and context-aware (ctx= 5) setups. N/A: not applicable as {\towerinst} is not trained on Arabic. \textbf{Bold} highlights the best value per row. The random chance performance on discourse phenomena varies depending on the number of elements in the list of ambiguous words, which differs across languages.}
\label{ted_results}
\end{table*}

\begin{table*}
\centering
\tiny
\setlength{\tabcolsep}{4pt}
\begin{tabular}{llcccccccccccccccccc}
\toprule
& \multirow{3}{*}{\textbf{Metric}} & \multicolumn{2}{c}{\textbf{nllb}} & \multicolumn{4}{c}{\textbf{tower}} & \multicolumn{4}{c}{\textbf{euro}}  & \multicolumn{4}{c}{\textbf{gemma}}  & \multicolumn{4}{c}{\textbf{qwen}} \\
  & & \multicolumn{2}{c}{\textbf{ctx= 0}} & \multicolumn{2}{c}{\textbf{ctx= 0}} & \multicolumn{2}{c}{\textbf{ctx= 5}} & \multicolumn{2}{c}{\textbf{ctx= 0}} & \multicolumn{2}{c}{\textbf{ctx= 5}} & \multicolumn{2}{c}{\textbf{ctx= 0}} & \multicolumn{2}{c}{\textbf{ctx= 5}} & \multicolumn{2}{c}{\textbf{ctx= 0}} & \multicolumn{2}{c}{\textbf{ctx= 5}} \\
& & \textbf{G} & \textbf{Q} & \textbf{G} & \textbf{Q} & \textbf{G} & \textbf{Q} & \textbf{G} & \textbf{Q} & \textbf{G} & \textbf{Q} & \textbf{G} & \textbf{Q} & \textbf{G} & \textbf{Q} & \textbf{G} & \textbf{Q} & \textbf{G} & \textbf{Q} \\
\midrule
\multirow{8}{*}{\textbf{EN-PT}} & \BLEU & 33.2 & 35.2 & 24.7 & 35.5 & 25.8 & 35.6 & 26.7 & 38.9 & 26.4 & 39.3 & 37.7 & 39.3 & 39.2 & \textbf{40.5} & 36.5 & 38.0 & 37.4 & 38.6 \\
& \COMET & 78.8 & 79.5 & 79.3 & 83.3 & 79.2 & 83.2 & 78.8 & 82.8 & 78.1 & 83.2 & 82.4 & 83.8 & 83.1 & 84.0 & 83.5 & 83.8 & 83.9 & \textbf{84.2} \\
& \docCOMET & 76.8 & 77.6 & 77.1 & 81.2 & 77.3 & 81.4 & 75.9 & 80.2 & 75.3 & 80.9 & 80.1 & 81.7 & 80.8 & 81.8 & 81.3 & 81.7 & 81.8 & \textbf{82.2} \\
& \COMETQE & 75.7 & 76.5 & 75.0 & \textbf{79.8} & 73.7 & 78.4 & 74.8 & 79.7 & 73.5 & 79.1 & 77.9 & 79.6 & 78.1 & 79.2 & 79.2 & 79.5 & 79.2 & 79.5 \\
& \docCOMETQE & 67.5 & 68.5 & 66.9 & \textbf{73.2} & 66.3 & 72.4 & 66.0 & 73.1 & 64.7 & 73.0 & 71.1 & 72.7 & 72.1 & 72.9 & 72.2 & 72.9 & 72.7 & 73.1 \\
& L. repetition & 77.0 & 76.0 & 75.0 & 79.0 & 78.0 & 83.0 & 79.0 & 82.0 & 79.0 & 84.0 & 79.0 & 80.0 & 84.0 & 84.0 & 80.0 & 83.0 & 85.0 & \textbf{85.0} \\
& Formality & 58.0 & 62.0 & 45.0 & 58.0 & 47.0 & 61.0 & 57.0 & \textbf{66.0} & 58.0 & \textbf{66.0} & 55.0 & 65.0 & 62.0 & 64.0 & 65.0 & \textbf{65.0} & 65.0 & 63.0 \\
& Pronouns & 49.0 & 50.0 & 20.0 & 35.0 & 42.0 & 49.0 & 45.0 & 45.0 & 46.0 & \textbf{56.0} & 47.0 & 48.0 & 46.0 & 45.0 & 49.0 & 44.0 & 44.0 & 47.0 \\
\midrule
\multirow{8}{*}{\textbf{EN-DE}} & \BLEU & 05.1 & 05.2 & 04.2 & 05.5 & 07.9 & \textbf{12.6} & 03.6 & 05.3 & 03.6 & 04.8 & 05.3 & 05.2 & 04.9 & 05.3 & 04.7 & 05.0 & 04.9 & 05.1 \\
& \COMET & 48.1 & 47.5 & 48.2 & 50.4 & 58.2 & \textbf{63.0} & 48.3 & 50.8 & 48.2 & 50.8 & 50.3 & 50.8 & 47.9 & 49.7 & 49.9 & 50.4 & 50.4 & 50.6 \\
& \docCOMET & 46.0 & 45.6 & 46.1 & 47.9 & 57.0 & \textbf{61.6} & 45.8 & 47.9 & 45.9 & 48.2 & 47.6 & 48.3 & 45.2 & 47.4 & 47.5 & 47.8 & 47.9 & 48.1 \\
& \COMETQE & 77.1 & 76.9 & 74.5 & \textbf{80.4} & 58.6 & 65.2 & 74.0 & 80.2 & 67.4 & 75.9 & 77.4 & 80.1 & 70.9 & 77.1 & 79.6 & \textbf{80.4} & 79.6 & 80.2 \\
& \docCOMETQE & 69.7 & 69.4 & 67.0 & 74.2 & 59.9 & 68.8 & 64.3 & 74.2 & 61.5 & 71.3 & 71.7 & 74.0 & 67.0 & 72.4 & 73.2 & \textbf{74.5} & 73.7 & \textbf{74.5} \\
& L. repetition & 26.0 & 27.0 & 33.0 & 29.0 & 22.0 & 23.0 & 32.0 & 29.0 & 30.0 & 29.0 & 30.0 & 30.0 & \textbf{34.0} & 33.0 & 28.0 & 28.0 & 29.0 & 28.0 \\
& Formality & 23.0 & 23.0 & 22.0 & 24.0 & 36.0 & \textbf{37.0} & 23.0 & 25.0 & 24.0 & 23.0 & 18.0 & 24.0 & 15.0 & 21.0 & 22.0 & 22.0 & 22.0 & 22.0 \\
& Pronouns & 26.0 & 23.0 & 19.0 & 28.0 & 22.0 & 14.0 & 22.0 & 26.0 & 26.0 & 23.0 & 27.0 & 27.0 & 25.0 & 26.0 & 28.0 & 27.0 & 26.0 & \textbf{29.0} \\
\midrule
\multirow{9}{*}{\textbf{EN-FR}} & \BLEU & 23.3 & 32.8 & 24.5 & 35.3 & 25.7 & 36.9 & 24.2 & 36.5 & 23.3 & 32.8 & 36.0 & 38.0 & 36.2 & \textbf{38.8} & 33.8 & 35.0 & 34.9 & 36.3 \\
& \COMET & 74.0 & 79.7 & 77.0 & 81.2 & 76.6 & 81.3 & 74.6 & 80.2 & 74.0 & 79.7 & 79.4 & 81.2 & 78.3 & 81.0 & 80.4 & 80.8 & 81.2 & \textbf{81.7} \\
& \docCOMET & 72.7 & 78.0 & 75.3 & 79.6 & 75.6 & 80.1 & 72.5 & 78.3 & 72.7 & 78.0 & 77.9 & 79.8 & 77.0 & 79.6 & 78.9 & 79.2 & 79.7 & \textbf{80.3} \\
& \COMETQE & 74.0 & 80.1 & 77.1 & \textbf{81.8} & 75.5 & 81.1 & 75.5 & 81.5 & 74.0 & 80.1 & 79.6 & 81.3 & 78.0 & 80.7 & 81.3 & 81.8 & 81.2 & 81.7 \\
& \docCOMETQE & 65.1 & 73.5 & 68.2 & 74.7 & 68.3 & 74.6 & 64.8 & 73.9 & 65.1 & 73.5 & 72.8 & 74.1 & 72.4 & 74.0 & 74.0 & 74.5 & 74.3 & \textbf{75.0} \\
& L. repetition & 66.0 & 72.0 & 69.0 & 75.0 & 72.0 & \textbf{80.0} & 65.0 & 73.0 & 66.0 & 72.0 & 73.0 & 75.0 & 73.0 & 77.0 & 73.0 & 74.0 & 79.0 & 78.0 \\
& Formality & 49.0 & 58.0 & 53.0 & 57.0 & 57.0 & 61.0 & 50.0 & \textbf{61.0} & 49.0 & 58.0 & 56.0 & 64.0 & 57.0 & \textbf{68.0} & 61.0 & 62.0 & 63.0 & 65.0 \\
& Pronouns & 47.0 & 53.0 & 47.0 & 47.0 & 50.0 & \textbf{60.0} & 44.0 & 50.0 & 47.0 & 53.0 & 51.0 & 53.0 & 57.0 & \textbf{60.0} & 50.0 & 49.0 & 58.0 & \textbf{60.0} \\
& Verb form & 37.0 & \textbf{49.0} & 27.0 & 43.0 & 30.0 & 45.0 & 35.0 & \textbf{49.0} & 37.0 & \textbf{49.0} & 47.0 & 48.0 & 44.0 & 47.0 & 45.0 & 46.0 & 45.0 & 45.0 \\
\midrule
\multirow{8}{*}{\textbf{EN-RU}} & \BLEU & 20.6 & 20.7 & 14.8 & 22.5 & 13.7 & 23.2 & 14.7 & 22.3 & 15.1 & 23.5 & 22.3 & 23.5 & 23.8 & \textbf{24.6} & 20.0 & 21.6 & 22.1 & 23.2 \\
& \COMET & 76.1 & 76.2 & 76.2 & 81.2 & 73.8 & 81.5 & 76.4 & 81.1 & 76.0 & 81.6 & 81.1 & 82.0 & 81.0 & \textbf{82.1} & 80.2 & 80.7 & 81.6 & 81.9 \\
& \docCOMET & 74.0 & 74.1 & 73.7 & 78.6 & 72.5 & 79.6 & 73.9 & 78.6 & 74.0 & 79.4 & 78.4 & 79.6 & 78.7 & \textbf{79.8} & 77.6 & 78.2 & 79.3 & 79.7 \\
& \COMETQE & 75.8 & 75.5 & 71.7 & 79.0 & 67.5 & 77.5 & 72.6 & 79.3 & 72.2 & 78.8 & 78.7 & 79.4 & 77.5 & 78.7 & 78.6 & \textbf{79.5} & 78.7 & 79.2 \\
& \docCOMETQE & 67.0 & 67.0 & 61.7 & 71.6 & 59.5 & 70.6 & 62.5 & 71.7 & 63.6 & 71.5 & 71.6 & \textbf{72.2} & 70.6 & 71.8 & 70.7 & 72.1 & 71.3 & \textbf{72.2} \\
& L. repetition & 63.0 & 53.0 & 50.0 & 63.0 & 65.0 & 72.0 & 42.0 & 60.0 & 64.0 & \textbf{75.0} & 46.0 & 49.0 & 72.0 & 73.0 & 48.0 & 48.0 & 72.0 & \textbf{75.0} \\
& Formality & 47.0 & 48.0 & 39.0 & 48.0 & 34.0 & \textbf{55.0} & 43.0 & 53.0 & 49.0 & 52.0 & 48.0 & 51.0 & 51.0 & 53.0 & 41.0 & 44.0 & 48.0 & 30.0 \\
& Verb form & 32.0 & 34.0 & 26.0 & 36.0 & 21.0 & \textbf{38.0} & 22.0 & 37.0 & 28.0 & 37.0 & 35.0 & \textbf{38.0} & 33.0 & \textbf{38.0} & 28.0 & 33.0 & 32.0 & 33.0 \\
\midrule
\multirow{7}{*}{\textbf{EN-AR}} & \BLEU & 17.5 & 16.9 & N/A & N/A & N/A & N/A & 00.3 & 00.5 & 10.3 & \textbf{20.6} & 00.4 & 00.5 & 00.4 & 00.5 & 00.4 & 00.4 & 00.5 & 00.5 \\
& \COMET & 77.8 & 77.1 & N/A & N/A & N/A & N/A & 50.1 & 52.4 & 75.3 & \textbf{81.7} & 53.4 & 53.7 & 50.6 & 52.3 & 50.7 & 52.5 & 50.7 & 52.4 \\
& \docCOMET & 75.2 & 74.3 & N/A & N/A & N/A & N/A & 46.6 & 48.7 & 72.7 & \textbf{79.2} & 49.5 & 50.0 & 47.2 & 49.0 & 47.2 & 48.9 & 47.4 & 49.0 \\
& \COMETQE & 72.7 & 71.0 & N/A & N/A & N/A & N/A & 65.7 & 74.9 & 66.8 & 75.1 & 74.6 & \textbf{76.1} & 69.1 & 72.9 & 70.1 & 74.8 & 69.7 & 74.4 \\
& \docCOMETQE & 64.7 & 63.1 & N/A & N/A & N/A & N/A & 55.3 & 68.1 & 57.5 & 67.9 & 68.0 & \textbf{69.4} & 65.4 & 68.0 & 61.6 & 67.3 & 61.6 & 67.2 \\
& L. repetition & 63.0 & 60.0 & N/A & N/A & N/A & N/A & 33.0 & 60.0 & 58.0 & \textbf{75.0} & 50.0 & 54.0 & 51.0 & 59.0 & 53.0 & 60.0 & 62.0 & 69.0 \\
& Pronouns & 55.0 & 43.0 & N/A & N/A & N/A & N/A & 36.0 & 44.0 & 36.0 & 54.0 & 39.0 & \textbf{56.0} & 41.0 & 41.0 & 49.0 & 53.0 & 51.0 & \textbf{56.0} \\
\midrule
\multirow{7}{*}{\textbf{EN-KO}} & \BLEU & 22.2 & 22.1 & 16.2 & 25.2 & 12.0 & 20.1 & 15.0 & 25.8 & 17.0 & 28.6 & 25.3 & 27.6 & 27.8 & \textbf{30.6} & 21.6 & 24.9 & 23.5 & 26.7 \\
& \COMET & 80.1 & 79.6 & 80.8 & 85.0 & 77.4 & 83.4 & 79.2 & 84.6 & 79.9 & 85.7 & 83.8 & \textbf{86.5} & 84.9 & 86.3 & 83.9 & 85.5 & 84.8 & 86.2 \\
& \docCOMET & 77.2 & 76.7 & 77.2 & 81.7 & 75.1 & 80.9 & 75.8 & 81.5 & 76.9 & 82.8 & 80.1 & 82.0 & 81.7 & \textbf{83.3} & 80.3 & 81.9 & 81.6 & 83.0 \\
& \COMETQE & 79.4 & 78.7 & 75.9 & 81.9 & 69.5 & 77.9 & 75.1 & 81.9 & 75.4 & 82.1 & 79.9 & 81.9 & 80.4 & 82.3 & 80.4 & \textbf{82.7} & 80.6 & 82.5 \\
& \docCOMETQE & 70.2 & 70.3 & 67.7 & 75.2 & 63.7 & 73.6 & 65.3 & 74.6 & 66.6 & 75.2 & 73.4 & 75.2 & 74.7 & 75.8 & 72.8 & 75.8 & 73.2 & \textbf{76.0} \\
& L. repetition & 44.0 & 32.0 & 22.0 & 37.0 & 34.0 & 43.0 & 23.0 & 41.0 & 38.0 & 57.0 & 39.0 & 44.0 & 55.0 & \textbf{62.0} & 32.0 & 35.0 & 48.0 & 50.0 \\
& Formality & 27.0 & 27.0 & 10.0 & 30.0 & 25.0 & 32.0 & 09.0 & 22.0 & 09.0 & 34.0 & 44.0 & 38.0 & 33.0 & \textbf{48.0} & 17.0 & 07.0 & 40.0 & 31.0 \\
\bottomrule
\end{tabular}
\caption{Detailed language-specific translation and discourse phenomena performance of all models (nllb={\nllb}, tower={\towerinst}, euro={\eurollm}, gemma={\gemmathree}, qwen={\qwenthree}) using greedy (G) and quality-aware decoding (Q) on \textbf{WMT24++ dataset} in both sentence-level (ctx= 0) and context-aware (ctx= 5) setups. N/A: not applicable as {\towerinst} is not trained on Arabic. \textbf{Bold} highlights the best value per row.}
\label{wmt_results}
\end{table*}
\begin{figure*}
\centering%
\centering%
\small
\begin{Verbatim}[frame=single, fontsize=\small, breaklines=true, breakanywhere=true, commandchars=\\\{\}]
\textbf{Translate the following <src_lang> source text to <tgt_lang>.}
\textbf{<src_lang>:} <src context 1> <src context 2> <src context 3> <src context 4> <src context 5> <src_sentence>
\textbf{<tgt_lang>}: <tgt context 1> <tgt context 2> <tgt context 3> <tgt context 4> <tgt context 5>
\end{Verbatim}
\caption{{\towerinst} prompt format.}
\label{tower-prompt-format}
\end{figure*}

\begin{figure*}
\centering%
\small
\begin{Verbatim}[frame=single, fontsize=\small, breaklines=true, breakanywhere=true,  commandchars=\\\{\}]
\textbf{<src_lang>:} <src context 1> \textbf{<tgt_lang>:} <tgt context 1>
\textbf{<src_lang>:} <src context 2> \textbf{<tgt_lang>:} <tgt context 2>
\textbf{<src_lang>:} <src context 3> \textbf{<tgt_lang>:} <tgt context 3>
\textbf{<src_lang>:} <src context 4> \textbf{<tgt_lang>:} <tgt context 4>
\textbf{<src_lang>:} <src context 5> \textbf{<tgt_lang>:} <tgt context 5>
\textbf{Given the provided parallel sentence pairs, translate the following <src_lang> sentence to <tgt_lang>.}
\textbf{<src_lang>:} <src sentence> \textbf{<tgt_lang>:} 
\end{Verbatim}
\caption{{\eurollm} prompt format.}
\label{eurollm-prompt-format}
\end{figure*}

\begin{figure*}
\centering%
\small
\begin{Verbatim}[frame=single, fontsize=\small, breaklines=true, breakanywhere=true,  commandchars=\\\{\}]
\textbf{<src_lang>:} <src context 1> \textbf{<tgt_lang>:} <tgt context 1>
\textbf{<src_lang>:} <src context 2> \textbf{<tgt_lang>:} <tgt context 2>
\textbf{<src_lang>:} <src context 3> \textbf{<tgt_lang>:} <tgt context 3>
\textbf{<src_lang>:} <src context 4> \textbf{<tgt_lang>:} <tgt context 4>
\textbf{<src_lang>:} <src context 5> \textbf{<tgt_lang>:} <tgt context 5>
\textbf{Given the provided parallel sentence pairs, translate the following <src_lang> sentence to <tgt_lang>. Do not give an explanation of your translation.}
\textbf{<src_lang>:} <src sentence> \textbf{<tgt_lang>:} 
\end{Verbatim}
\caption{{\gemmathree} and {\qwenthree} prompt format.}
\label{gemma-qwen-prompt-format}
\end{figure*}
\begin{table*}
\small
\begin{tabular}{|p{0.95\textwidth}|}
\hline
We present the participants with 25 samples including the following data:
\begin{itemize}
\item The \textbf{source context} which was given to the translation model, which are (up to 5) previous sentences in the source document.
\item The English source sentence.
\item The \textbf{output context} which was given to the translation model, which are (up to 5) previous sentences in the output document.
\item \textbf{output 1}: the output of the first system
\item \textbf{output 2}: the output of the second system
\end{itemize}

Annotators are asked to assess the following:

\begin{itemize}
    \item \textbf{Semantic difference}: Rate the semantic difference of the two outputs on a scale of 1 to 5, ignoring differences in wording. Consider whether they convey the same meaning.
    \begin{itemize}
        \item 1: the two sentences convey the same meaning.
        \item 5: the two sentences convey completely different meanings.
    \end{itemize}

    \item \textbf{Pronoun resolution}: Does the source sentence contain an ambiguous pronoun (a pronoun whose referent is unclear or not explicitly mentioned), and what is it?
    \begin{itemize}
        \item If yes, is it correctly translated in output 1?
        \item If yes, is it correctly translated in output 2?
    \end{itemize}
    \item \textbf{Lexical repetition}: Does the source sentence contain an entity (e.g., noun, occupation) previously mentioned in the source context, and what is the entity?
    \begin{itemize}
        \item If yes, is it translated consistently with its previous translation in the output context in output 1?
        \item If yes, is it translated consistently with its previous translation in the output context in output 2?
    \end{itemize}
    \item \textbf{Formality}: Does the source sentence exhibit a formality phenomenon (e.g., addressing someone formally or expressing respect), and what is the word that exhibits the phenomenon? 
    \begin{itemize}
        \item If yes, is it handled in the output 1?
        \item If yes, is it handled in the output 2?
    \end{itemize}
    \item \textbf{Verb form}: Does the source sentence contain an ambiguous verb that can have different forms depending on the gender or formality level of the subject, and what is the verb?
    \begin{itemize}
        \item If yes, is it correctly translated in output 1?
        \item If yes, is it correctly translated in output 2?
    \end{itemize}

    \item \textbf{General comment (optional)}: Provide comments or observations about the two outputs. Highlight strengths, weaknesses, or notable phenomena (e.g., mistranslation, cultural adaptation, or syntactic errors). Please also highlight other linguistic phenomena we may have missed in the categories provided.
    \item \textbf{Preference}: Which output do you prefer? (output 1, output 2, equally good, equally bad)
\end{itemize}\\
\hline
\end{tabular}
\caption{Human assessment details.}
\label{Human_assessment_details}
\end{table*}

\section{Qualitative Examples}
\label{qualitative_examples}
\begin{figure*}
\centering
\small
\includegraphics[width=\textwidth]{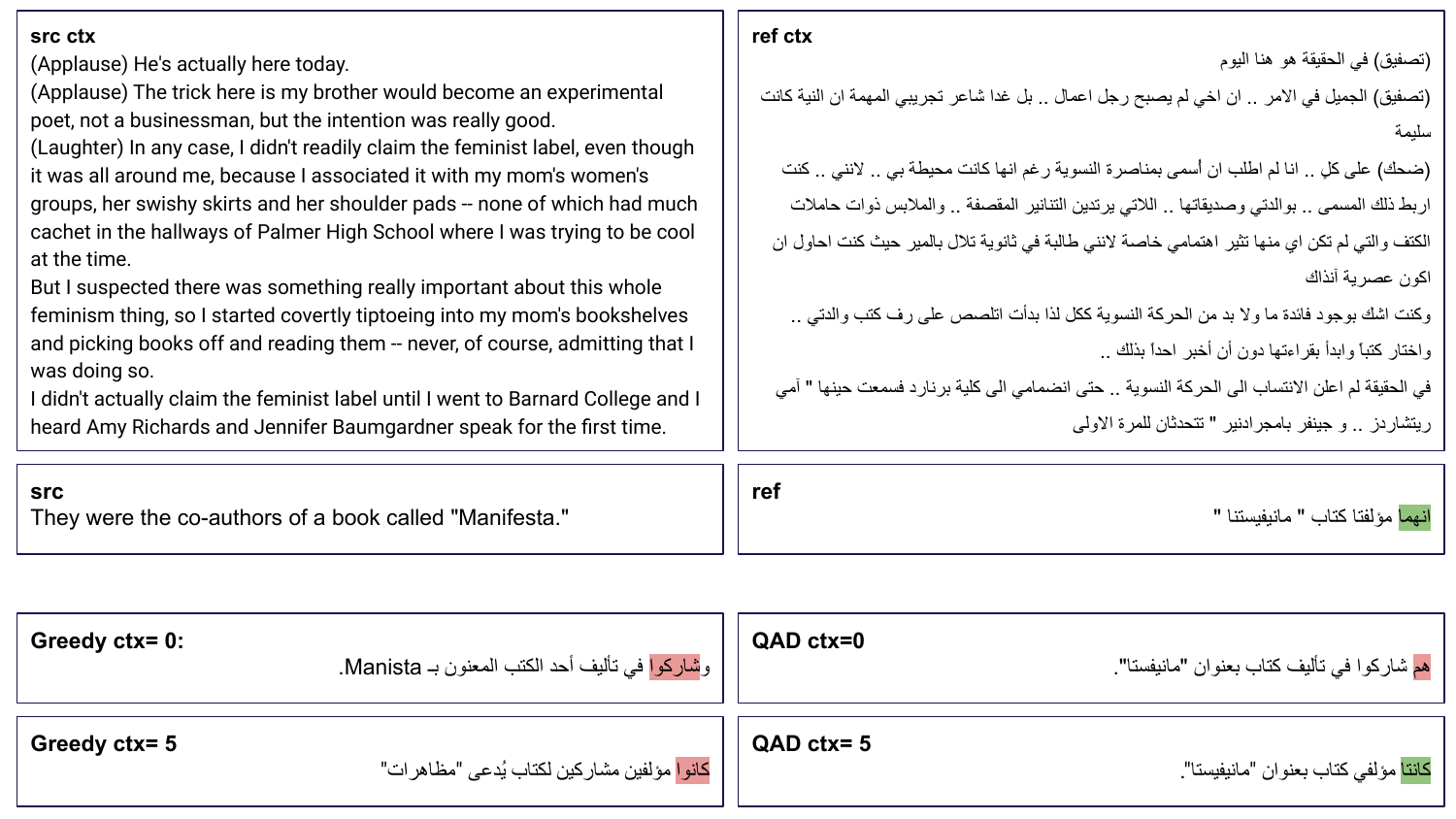}
\caption{Dual pronouns in Arabic}
\label{example1}
\end{figure*}

\begin{figure*}
\centering
\small
\includegraphics[width=\textwidth]{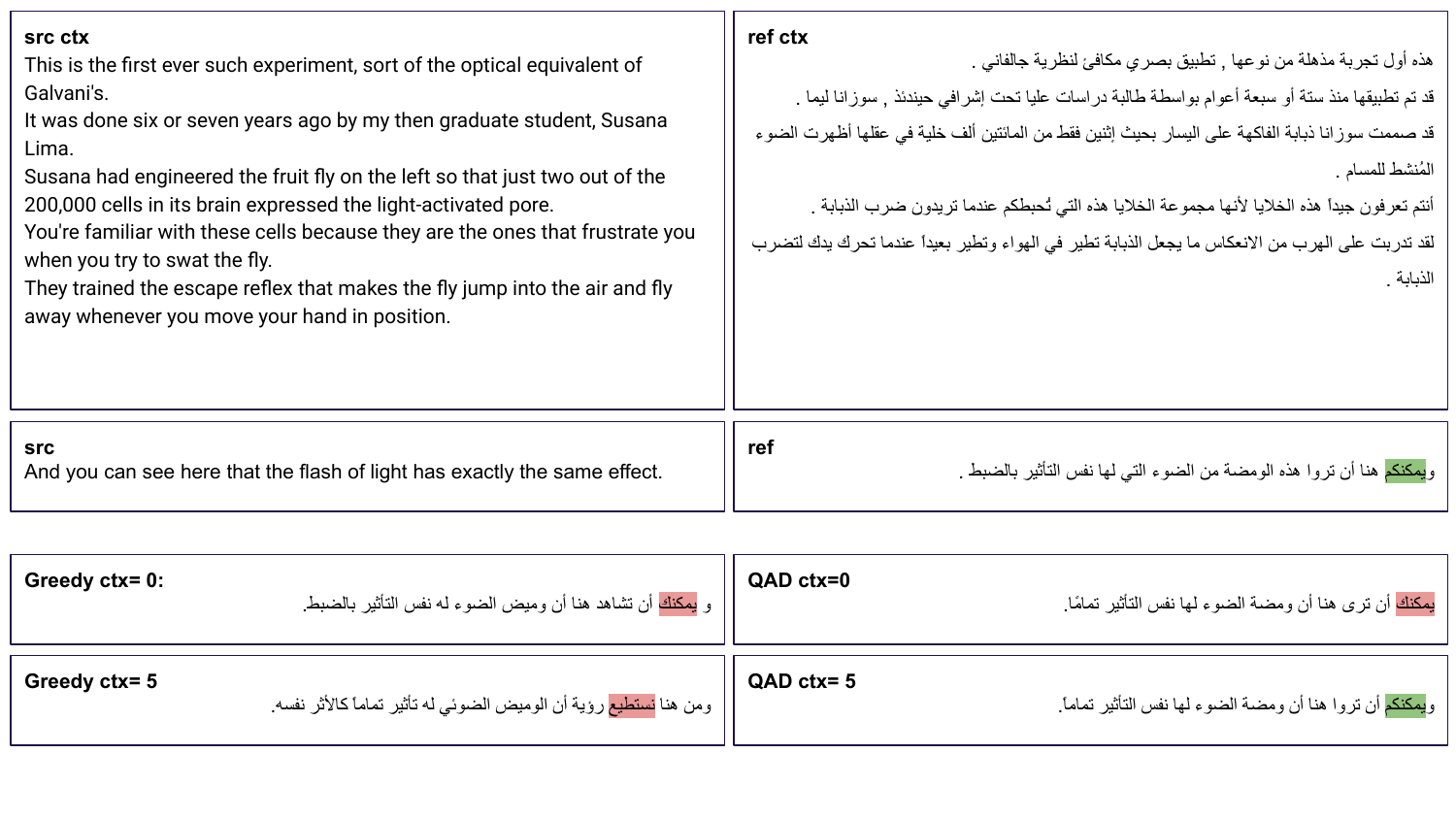}  
\caption{Plural pronouns in Arabic.}
\label{example2}
\end{figure*}

\begin{figure*}
\centering
\includegraphics[width=\textwidth]{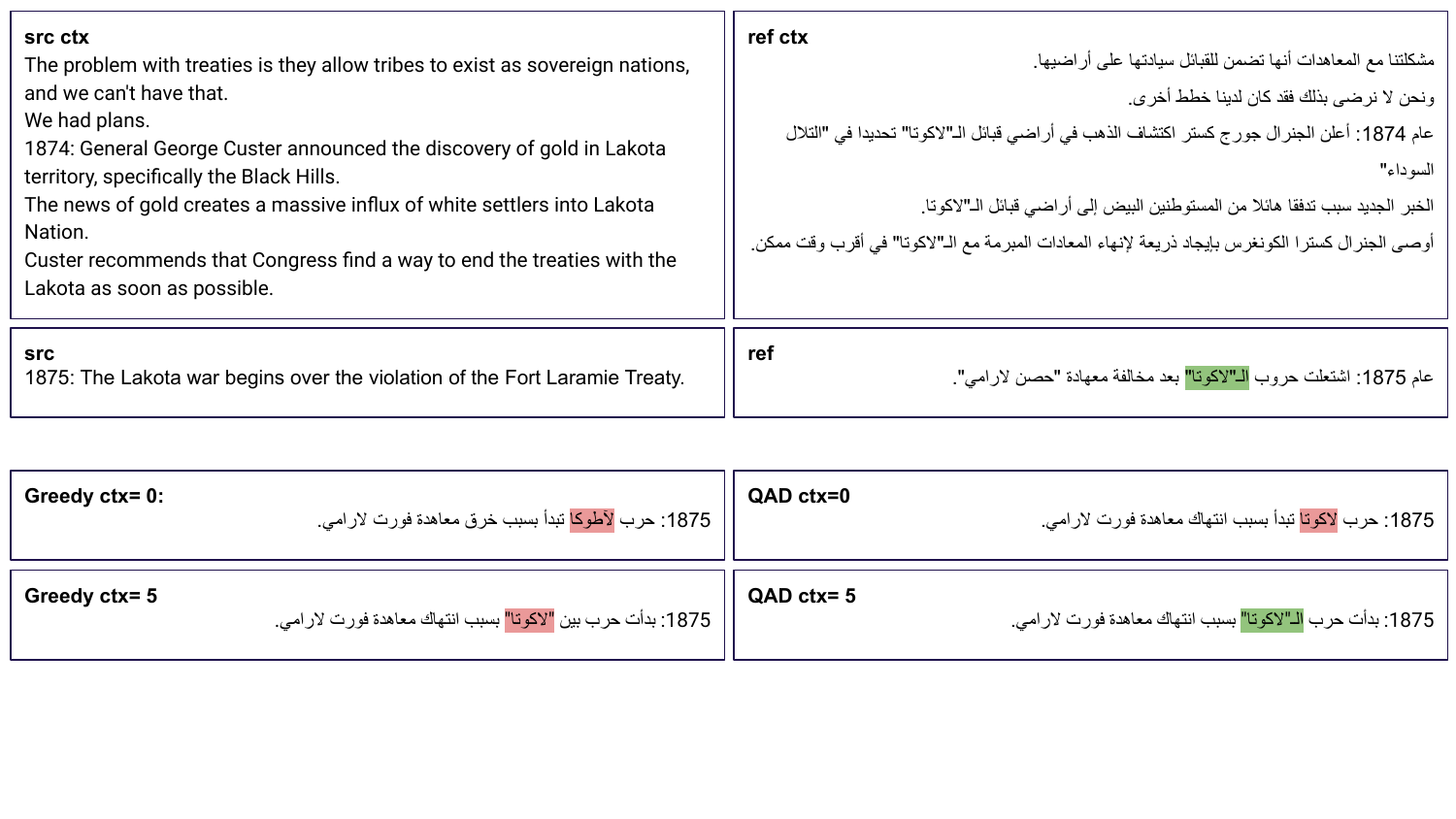}  
\caption{Lexical repetition in Arabic.}
\label{example3}
\end{figure*}

\begin{figure*}
\centering
\includegraphics[width=\textwidth]{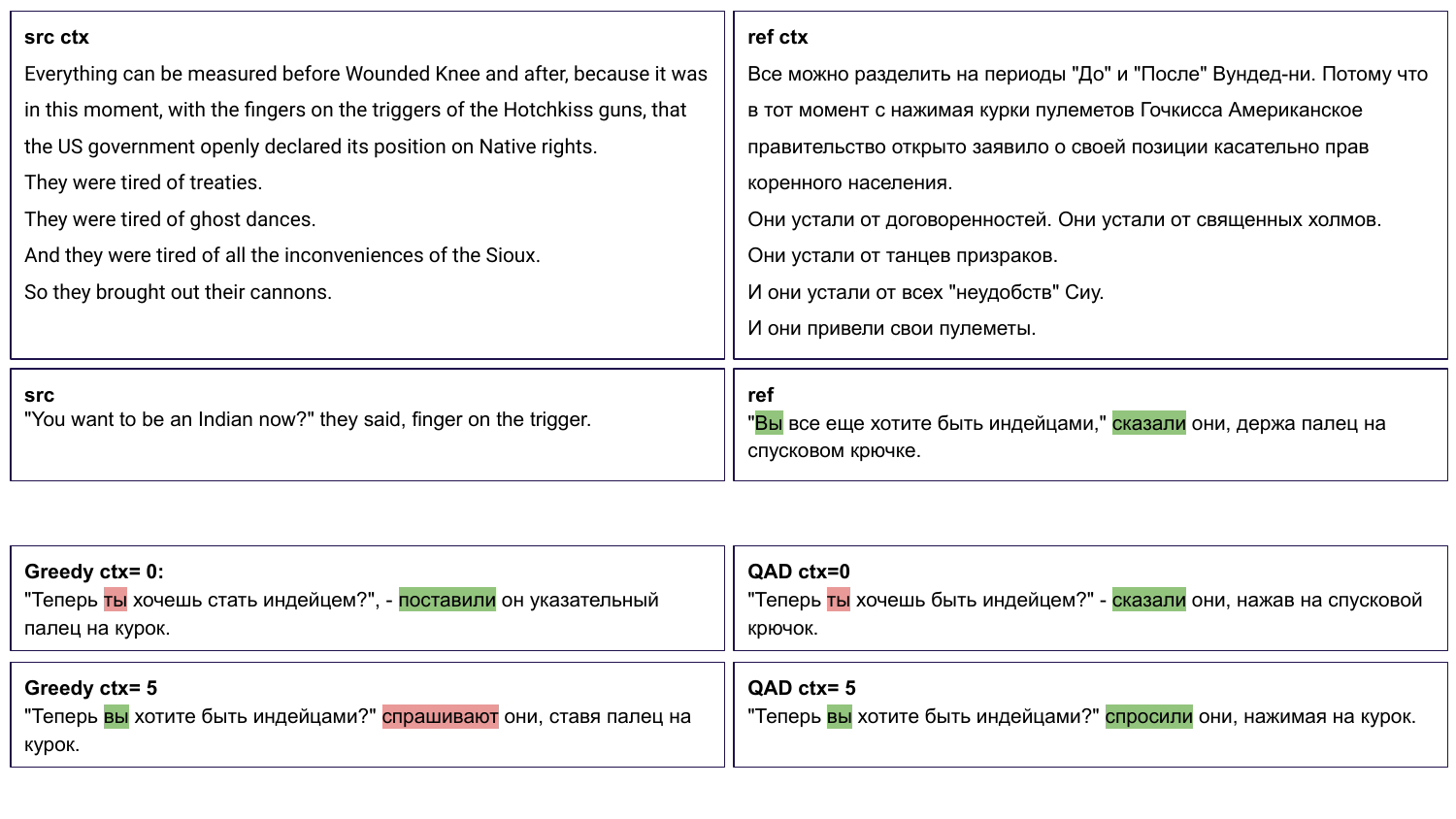}  
\caption{Plural pronoun and verb form in Russian.}
\label{example4}
\end{figure*}

\begin{figure*}
\centering
\includegraphics[width=\textwidth]{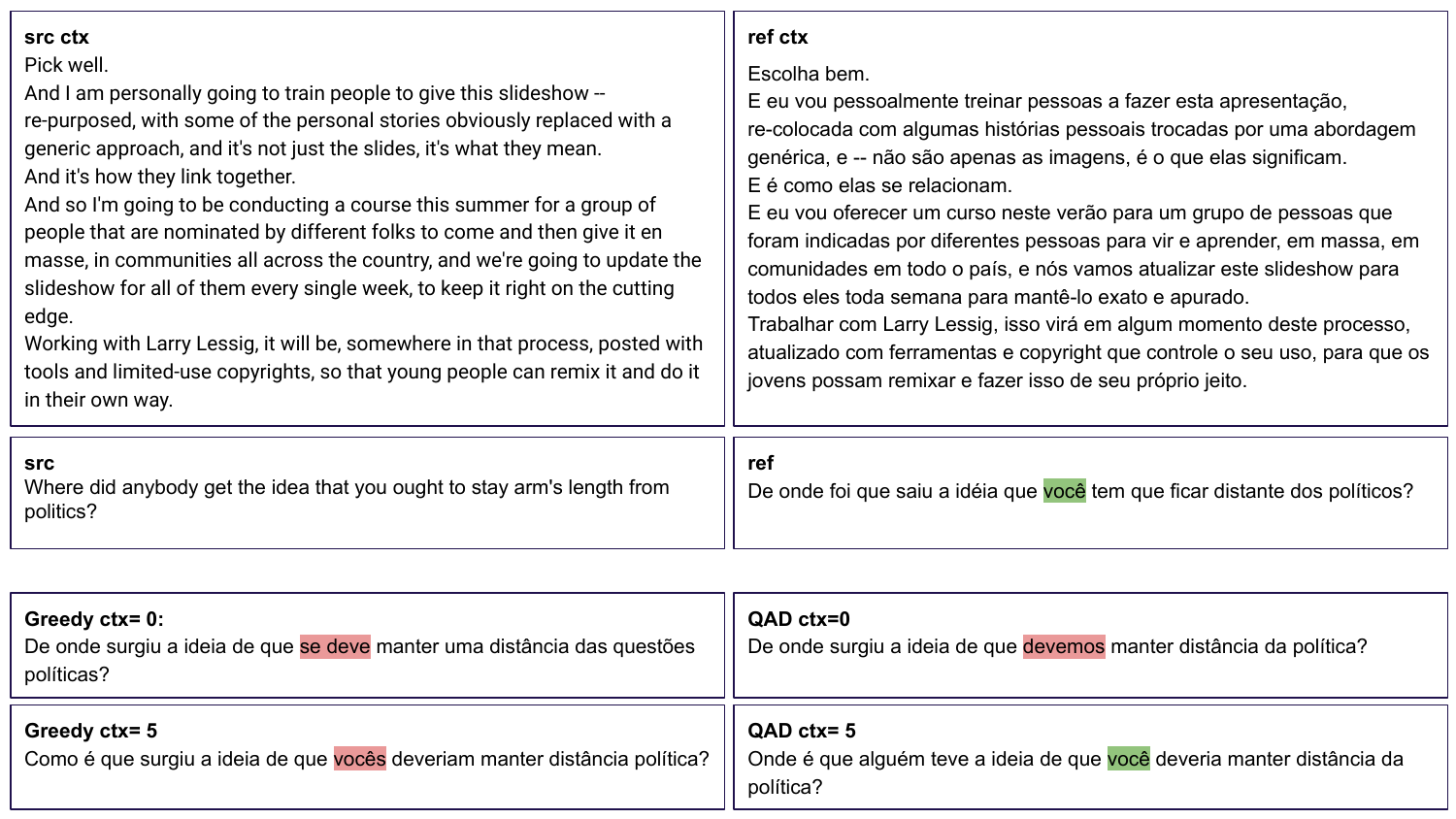}  
\caption{Formality and pronoun in Brazilian-Portuguese.}
\label{example5}
\end{figure*}

With the help of human annotators, we manually analyze the outputs of the sentence-level and context-aware greedy and QAD setups using {\eurollm} model for Arabic and {\towerinst} model for other languages to understand how QAD improves specific discourse phenomena. The analysis is performed on TED2020 dataset. We summarize our findings and present specific examples below:
\begin{itemize}
    \item \textbf{Dual-pronoun disambiguation:} using context-aware QAD, the reference of the pronoun \textit{they} to a dual subject in Arabic is correctly resolved, and the appropriate form is applied (\Cref{example1}).

    \item \textbf{Plural-pronoun disambiguation:} using context-aware QAD, the reference of the pronoun \textit{you} to a plural subject in Arabic is correctly resolved, and the appropriate form is applied (\Cref{example2}).

    \item \textbf{Lexical repetition:} using context-aware QAD, the entity \textit{Lakota} is translated consistently with the context in Arabic (\Cref{example3}). 

    \item \textbf{Plural-pronoun and past verb form:} using context-aware QAD, the reference of the pronoun \textit{you} to a plural subject in Russian is correctly resolved. Additionally, the correct past verb form \textit{said} is used. Context-aware QAD is the only setup where both phenomena are handled (\Cref{example4}).

    \item \textbf{Formality and pronoun:} using context-aware QAD, the correct form of the singular second person pronoun \textit{you} in Brazilian-Portuguese is used, it also maintains the correct formality level of explicitly mentioning the pronoun which conveys a conversational tone rather than an implicit pronoun which is more formal and detached (\Cref{example5}). 
     
\end{itemize}

\end{document}